\documentclass[11pt,letterpaper]{article}

\usepackage[margin=1in]{geometry}

\usepackage{stmaryrd}

\usepackage{amsmath}
\usepackage{amssymb}
\usepackage{amsfonts}
\usepackage{amsthm}
\usepackage{nicefrac}

\DeclareMathOperator{\vol}{Vol}

\newtheorem{theorem}{Theorem}[section]
\newtheorem{definition}{Definition}[section]
\newtheorem{proposition}{Proposition}[section]
\newtheorem{lemma}{Lemma}[section]
\newtheorem{remark}{Remark}[section]

\usepackage{graphicx}
\usepackage{booktabs}
\usepackage{multirow}
\usepackage{arydshln}
\usepackage{subcaption}

\newcommand{\bigtimes}{%
  \mathop{\scalebox{1.8}{$\displaystyle\times$}}\displaylimits
}

\usepackage{algorithm}
\usepackage{algpseudocode}

\usepackage{tikz}
\usepackage{pgfplots}
\usetikzlibrary{arrows.meta,positioning,calc}
\pgfplotsset{compat=1.18}

\usepackage{xcolor}
\usepackage{microtype}
\usepackage{comment}
\usepackage{url}

\usepackage[round]{natbib}
\usepackage[colorlinks=true,
            linkcolor=blue,
            citecolor=blue,
            urlcolor=blue]{hyperref}

\title{
    \textbf{Generalized Hyperrectangular Prediction Sets
    in Multivariate Conformal Prediction}
}

\author{
    Nabil Alami \qquad Souhaib Ben Taieb\\[0.5em]
    \small Computing and Mathematical Sciences Division\\
    \small MBZUAI, Abu Dhabi, UAE
}

\date{}

\begin{document}

\maketitle

\begin{abstract}
In multivariate conformal prediction, many nonconformity scores yield prediction sets with complex geometry (e.g., ellipsoids or density level sets) that are difficult to interpret and visualize. Hyperrectangles offer a practical alternative by providing one interval per output. We extend this notion to Cartesian products of arbitrary one-dimensional sets, which we call generalized hyperrectangles. We introduce a rectangularization framework that turns any continuous multivariate nonconformity score, or conditional density, into one yielding generalized hyperrectangular prediction sets with finite-sample coverage. We  establish key geometric properties of our rectangularization framework and provide useful intuitions on generalized hyperrectangles. We further establish a representation theorem showing that a broad class of hyperrectangular conformal methods based on max-aggregated coordinate-wise scores can be interpreted as rectangularizations of implicit conditional densities.
Through synthetic and real-world experiments, we quantify the efficiency cost of rectangularization, while highlighting its benefits, especially for practical applications of uncertainty quantification.

%Conformal prediction in multivariate settings often produces complex prediction sets (e.g., $\ell_p$ balls or density level sets) that are difficult to interpret and visualize. Hyperrectangles offer a practical alternative. Here, we extend this notion to Cartesian products of one-dimensional sets, which we call generalized hyperrectangles. We introduce a rectangularization framework that makes any multivariate nonconformity score yield  generalized hyperrectangular prediction sets with finite-sample coverage. We also provide an anlagous thing bur for densities.  Under suitable assumptions, we show how our approach unifies existing hyperrectangular conformal methods by showing how their prediction sets arise from a specific class of density-based scores. Through synthetic and real-world experiments, we quantify the efficiency cost of our rectangularization procedure, while demonstrating the practicality and benefits of using generalized hyperrectangular conformal prediction sets.

\end{abstract}

%\textcolor{blue}{say that multioutput is %kinda 'useless' as we cannot see it except write the set mathematically, however, it is not a reason to throw everything away: we will use these well-developed scores in multi-output CP to 'extract' from them score yielding hyperrectangular prediction sets.}%in the middle of the way, we will see that if can yield hyperrectangular prediction sets.

%\textcolor{blue}{say clearly that we define interpretability as the ability to visualize each coordinate-wise set. use unified notation of '\textbf{Cartesian Prediction sets}' (CPS).}

%\textcolor{blue}{state prop 3.3 of our apporach smaller than rectangular hull of the prediction set in the abstract?}

%\textcolor{blue}{very nice remark: from the tabme, if Gaussian Y | X or if we use mahalnbois score (i.e we take into account conditonal covariance), our ∞-rectangularization depends only on the diagonal variances: o many covariance structures collapse to the same rectan-gularized density!}

\section{Introduction}

Uncertainty quantification is a core challenge in modern statistical learning, and conformal prediction (CP) has emerged as a powerful framework for constructing prediction sets with finite-sample coverage under minimal assumptions~\citep{vovk2005algorithmic, papadopoulos2002inductive}. While CP is well understood for univariate outputs, there is growing interest in extending it to multivariate outputs.

\begin{figure}[ht]
    \centering
    \includegraphics[width=0.9\linewidth]{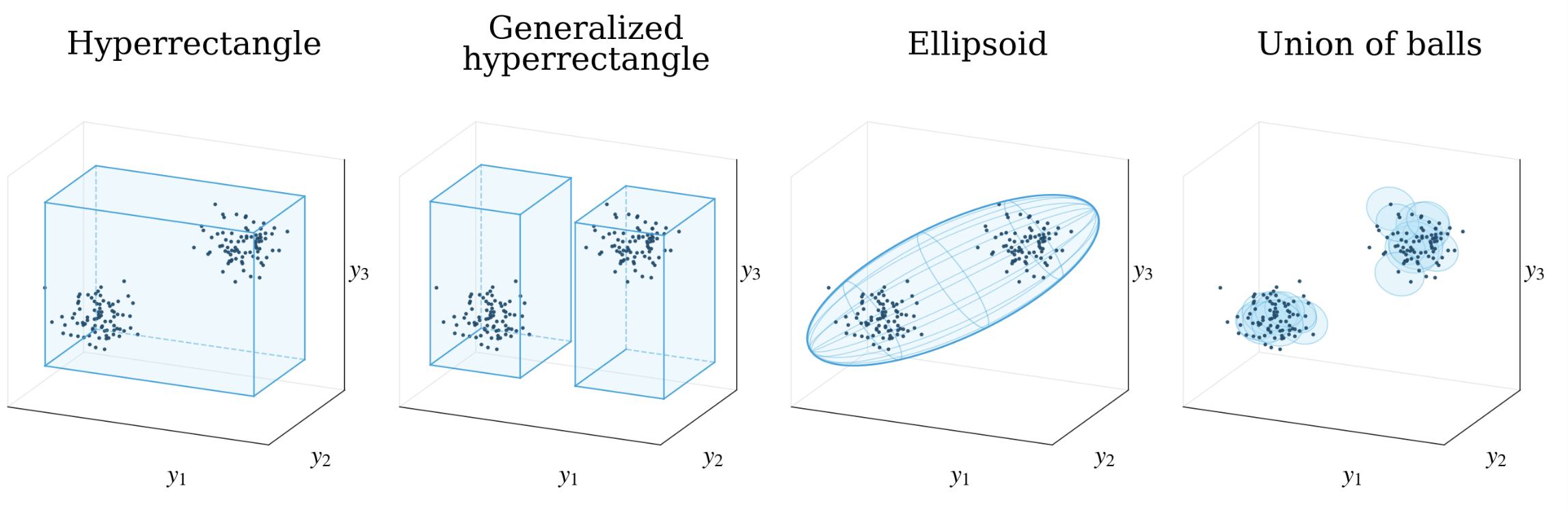}
    \caption{Multivariate CP sets: hyperrectangle, generalized hyperrectangle, ellipsoid~\citep{ellipse1},  and sampling-based set~\citep{wang2022probabilistic} for a bimodal data distribution. }

        \label{fig:introduction_figure_predsets}
\end{figure}

Recent multivariate CP methods exploit covariance structure~\citep{messoudi2022ellipsoidal, braun2025minimumvolumeconformalsets}, quantile regression~\citep{CQR, MCP_zhou2024conformalized}, generative models~\citep{wang2022probabilistic}, or conditional density estimates~\citep{izbicki2020cd, Dheur2025-unified}. These approaches can produce flexible prediction sets with small volume, but their geometry is often complex, in the sense that the sets are defined only implicitly, as sublevel sets of a multivariate score, and are difficult or impossible to visualize beyond two or three dimensions (Figure~\ref{fig:introduction_figure_predsets}). In this work, we call a prediction set \emph{interpretable} if it can be represented through one set per output coordinate, so that one can read off the plausible values of each output. This form of interpretability matters whenever practitioners must act on, or communicate, uncertainty about each output individually.\\

Hyperrectangular prediction sets, i.e., Cartesian products of intervals, are interpretable by construction (first panel of Figure~\ref{fig:introduction_figure_predsets}) and have a closed-form volume, although this type of geometry generally increases volume and thus reduces statistical efficiency. They were among the earliest constructions in multivariate CP~\citep{neeven2018conformal} and have since been extended in many directions~\citep{hyperectangle_CP,MCP_zhou2024conformalized,fan2025interpretablemultivariateconformalprediction,dcp_mt}. Existing hyperrectangular methods, however, are designed case by case, typically by aggregating univariate scores through a maximum or by calibrating each coordinate separately. This raises two questions. First, can a  hyperrectangular prediction set with finite-sample coverage be constructed from any existing multivariate nonconformity score? Second, can existing hyperrectangular methods be understood through a common  view?

To bridge this gap, this paper explores  both questions for a class of sets we call \emph{generalized hyperrectangles}: Cartesian products of arbitrary, possibly disconnected, measurable subsets of the real line (see the second panel of Figure~\ref{fig:introduction_figure_predsets}).

\paragraph{Summary of contributions.}
(i) We characterize the functions whose sublevel sets are generalized hyperrectangles (Section~\ref{sec:score_rectangularization}).
(ii) We introduce \emph{score rectangularization}, which turns scores for multivariate output into one whose conformal sets are generalized hyperrectangles with finite-sample coverage, and establish key theoretical properties (Sections~\ref{sec:score_rectangularization}--\ref{sec:density_rect}).
(iii) We prove a representation theorem showing that max-aggregated hyperrectangular methods arise from rectangularizing an implicit conditional density (Section~\ref{sec:rect_implicit}).
(iv) We quantify the volume cost of rectangularization and illustrate its benefits and interpretability on synthetic and real-world data (Section~\ref{sec:experiments}).

\section{Background and Notation}
\label{sec:background}

\textbf{Setup and notation.} For $t\in\mathbb{N}$, let $[t]:=\{1,\dots,t\}$. For a vector $y=(y_1,\dots,y_d)\in\mathbb{R}^d$ and $j\in[d]$, let $y_{-j}$ denote the vector of all coordinates of $y$ except $y_j$, and denote the Cartesian product of sets $A_1,\dots,A_d\subset\mathbb{R}$ by $\bigtimes_{j=1}^d A_j$. We consider a multi-output regression problem in which the goal is to predict an output vector $y\in\mathcal{Y}=\mathbb{R}^d$ from a feature vector $x\in\mathcal{X}\subseteq\mathbb{R}^M$, where $d,M\ge 2$. All random variables are defined on a common probability space $(\Omega,\mathcal{F},\mathbb{P})$. We assume a true joint distribution $P_{X,Y}$ on $\mathcal{X}\times\mathcal{Y}$ and access to a dataset $\mathcal{D}=\{(X_i,Y_i)\}_{i=1}^N$ with $(X_i,Y_i)\stackrel{\text{i.i.d.}}{\sim}P_{X,Y}$. For $x\in\mathcal{X}$, let $p_{Y|X}(y\mid x)$ denote the conditional density of $Y$ given $X=x$ with respect to the Lebesgue measure $\mathrm{Vol}(\cdot)$ on $\mathbb{R}^d$, and let $F_{Y|X}(y\mid x)$ denote the corresponding conditional distribution function.

\paragraph{Conformal Prediction.} CP is a widely used method for constructing prediction sets with finite-sample marginal coverage \citep{vovk2005algorithmic,lei2018distribution,angelopoulos2024theoretical}. We focus on Split-Conformal Prediction (SCP), where $\mathcal{D}$ is split into disjoint training and calibration sets $\mathcal{D}_{\mathrm{tr}}$ and $\mathcal{D}_{\mathrm{cal}}:=\{(X_i, Y_i)\}_{i=1}^n$ (after relabeling). We first train a model on $\mathcal{D}_{\text{tr}}$ to obtain a base predictor $\hat h$. Now consider any miscoverage level $\alpha\in(0,1)$. Given a test feature $X_{n+1}$ with unobserved response $Y_{n+1}$, where  $(X_{n+1},Y_{n+1})\sim P_{X,Y}$,  the goal of SCP is to construct a prediction set $\mathcal{C}_\alpha(X_{n+1})$  such that it contains $Y_{n+1}$ with probability at least $1-\alpha$.

This procedure requires a nonconformity score (NCS) function $$s:\mathcal{X} \times \mathcal{Y} \to \mathbb{R},$$ where lower scores indicate a better fit between the
feature vector $x$ and the response $y$ (for example, $s(x,y) = \| y - \hat h(x)\|_2$). We then compute the calibration scores $S_i := s(X_i, Y_i)$ for $i=1,\dots,n$, and define $\hat q_\alpha$ as the $\lceil (n+1)(1-\alpha)\rceil$-th smallest of $S_1,\dots,S_n$ (with $\hat q_\alpha:=+\infty$ if $\lceil (n+1)(1-\alpha)\rceil>n$). The SCP prediction set for $X_{n+1}$ is defined as 
\begin{equation}
\label{split-CP-set} 
\mathcal{C}_\alpha(X_{n+1};\mathcal{D}_{\mathrm{cal}}; s) := 
\{ y \in \mathcal{Y} : s(X_{n+1}, y) \le \hat q_\alpha \}.
\end{equation}

\begin{theorem}
\label{theorem:split-CPgaurantee} \citep{papadopoulos2002inductive}
%Conditionally on the training data $\mathcal{D}_{\text{tr}}$ used to constructthe NCS, 
Under exchangeability of $\mathcal{D}_{\text{cal}}$ and $(X_{n+1},Y_{n+1})$, the SCP
set \eqref{split-CP-set} satisfies the 
finite-sample coverage guarantee: \begin{equation}
\label{SCP_finite_sample_cov}
    \mathbb{P}(Y_{n+1}\in\mathcal{C}_\alpha(X_{n+1};\mathcal{D}_{\mathrm{cal}};s))\ge 1-\alpha.
\end{equation}
\end{theorem}

Throughout the remainder of the paper, we fix a calibration set $\mathcal{D}_{\mathrm{cal}}$ and suppress its dependence in the notation. For any feature vector $x\in\mathcal{X}$ and nonconformity score $s$, we will denote by $\mathcal{C}_\alpha(x;s)$ the prediction set defined in \eqref{split-CP-set}, with the quantile computed from the scores of $s$ on $\mathcal{D}_{\mathrm{cal}}$. The oracle prediction set will be denoted by $\mathcal{C}_\alpha^\star(x;s)$ when the empirical quantile $\hat q_\alpha$ is replaced by its population counterpart $q_\alpha$.

\section{Generalized Hyperrectangles and Related Work}
\label{sec:GH_related}

\paragraph{Prediction set geometry.}
Theorem~\ref{theorem:split-CPgaurantee} holds for any NCS: the choice of $s$ does not affect validity, but it entirely determines the geometry, and hence the volume, of $\mathcal{C}_\alpha(x;s)$. For instance, norm-based scores $s(x,y)=\|y-\hat h(x)\|$ yield convex prediction sets \citep{neeven2018conformal,messoudi2021copula,tumu2024multi}; in particular, the $\ell_\infty$ norm yields a \emph{hyperrectangle}, i.e., a set of the form $I_1\times\cdots\times I_d$ where each $I_j\subset\mathbb{R}$ is an interval. Hyperrectangular sets are attractive in multi-output CP because they provide one interval per output (first panel of Figure~\ref{fig:introduction_figure_predsets}). At the other end of the spectrum, density-based NCSs may yield flexible sets with irregular geometry and no simple analytic form \citep{Dheur2025-unified,wang2022probabilistic}.

\paragraph{Generalized hyperrectangles.}
We consider a broader class of geometries that extends hyperrectangles by allowing each coordinate set to be an arbitrary measurable subset of $\mathbb{R}$ rather than an interval.

\begin{definition}
\label{def:generalized_hyperr}
For $d\geq 1$, a generalized hyperrectangle (GH) is a set $\mathcal{H}\subseteq\mathbb{R}^d$ of the form
\[
\mathcal{H}=\bigtimes_{j=1}^d H_j,
\]
where $H_1,\ldots,H_d\subseteq\mathbb{R}$ are measurable sets, referred to as the coordinate sets of $\mathcal{H}$.
\end{definition}

If every $H_j$ is an interval, $\mathcal{H}$ is a hyperrectangle. If each $H_j$ is a finite union of intervals, $\mathcal{H}$ is a finite union of hyperrectangles. %Throughout, we write $\mathcal{C}_\alpha(x;s)$ for a generic prediction set, and $\mathcal{H}_\alpha(x;s)$ when the set is a GH.

\begin{figure}[htbp]
    \centering
    \begin{minipage}[b]{0.48\textwidth}
        \centering
        \includegraphics[width=\linewidth]{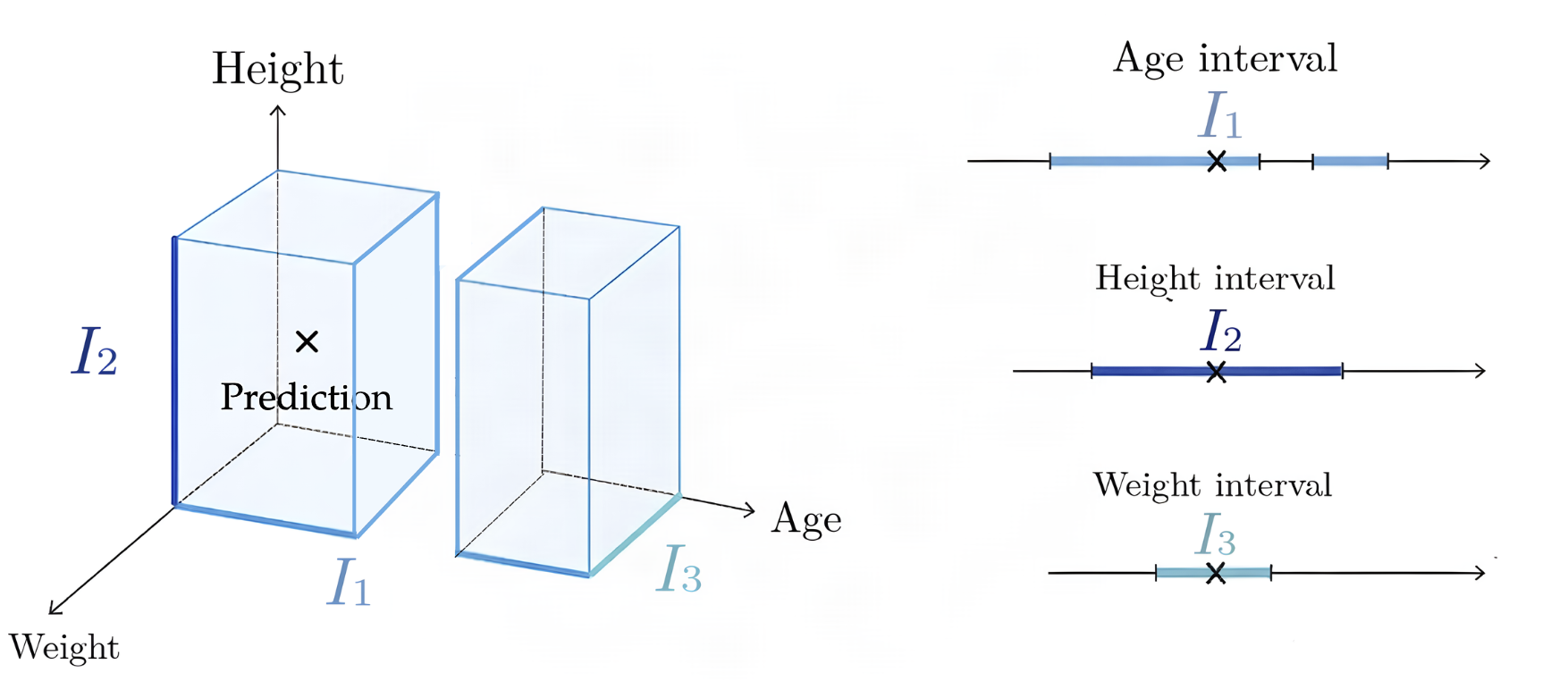}
        \caption{Illustration of a GH prediction set for a $3$-dimensional output.} %$Y:=(\text{Weight},\text{Age},\text{Height})$.}
        \label{fig:GH_example}
    \end{minipage}
    \hfill
    \begin{minipage}[b]{0.48\textwidth}
        \centering
        \includegraphics[width=\linewidth]{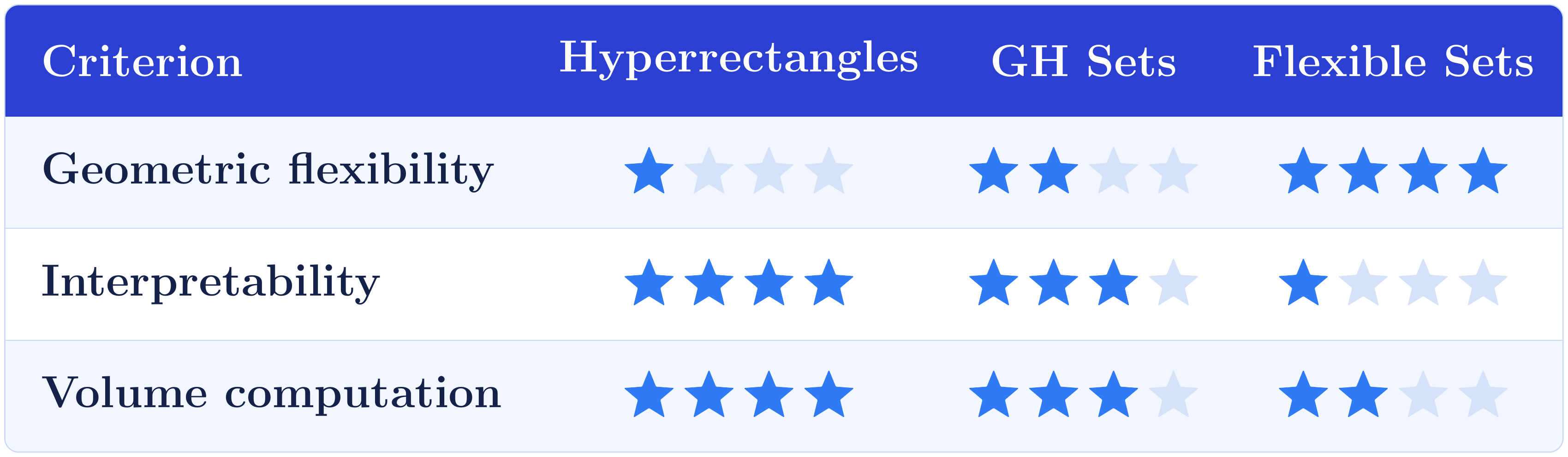}
        \caption{Qualitative comparison of prediction sets geometries.}
        \label{fig:sets_comparison}
    \end{minipage}
\end{figure}

\paragraph{Why GH prediction sets?}
GH sets offer three main advantages in multi-output CP:
\begin{itemize}
    \item[(a)]

\textbf{Flexibility.} Coordinate-wise sets may be disconnected, potentially resulting in substantially smaller volumes than hyperrectangles. (Figure~\ref{fig:GH_example}).

\item[(b)] \textbf{Interpretability.} A GH is fully specified by its one-dimensional coordinate sets $H_1,\ldots,H_d$ (Figure~\ref{fig:GH_example}), which can be inspected and visualized separately, unlike flexible sets such as ellipsoids or unions of balls (right panel of Figure~\ref{fig:introduction_figure_predsets}).

\item[(c)] \textbf{Tractable volume.} The volume of a GH is $\vol(\mathcal{H})=\prod_{j=1}^d |H_j|$, where $|\cdot|$ denotes the Lebesgue measure on $\mathbb{R}$, avoiding the Monte Carlo estimation required by more flexible sets \citep{Dheur2025-unified}.
\end{itemize}
The price of these advantages is that the Cartesian product structure may not exploit dependence across outputs, so GH sets can be larger than flexible sets; our experiments confirm this trade-off (Section~\ref{sec:experiments}).
For notation purposes, general prediction sets are denoted by $\mathcal{C}_\alpha(x; s)$, and generalized hyperrectangles (GH) by $\mathcal{H}_\alpha(x; s)$ to emphasize their geometry.

We illustrate qualitatively the differences of these prediction set geometries in Figure \ref{fig:sets_comparison}.

\paragraph{Max-score hyperrectangular methods.}
A common strategy for constructing hyperrectangular sets is to define univariate NCSs $s_1,\ldots,s_d:\mathcal{X}\times\mathbb{R}\to\mathbb{R}$, one per output, and aggregate them through a maximum. For a calibration sample $(X_i,Y_i)$ with $Y_i=(Y_{i1},\ldots,Y_{id})$, one computes $S_i:=s_{\max}(X_i,Y_i)=\max_{j\in[d]} s_j(X_i,Y_{ij})$, obtains $\hat q_\alpha$ from $\{S_i\}_{i=1}^n$, and defines
\begin{equation}
\label{eq:hyperrect_cp}
\begin{aligned}
\mathcal{H}_\alpha(x;s_{\max})
&=\Big\{y\in\mathbb{R}^d : \max_{j\in[d]} s_j(x,y_j)\le \hat q_\alpha\Big\}\\
&=\bigtimes_{j=1}^d\big\{y_j\in\mathbb{R}: s_j(x,y_j)\le \hat q_\alpha\big\}.
\end{aligned}
\end{equation}
Given a point predictor $\hat\mu:\mathcal{X}\to\mathcal{Y}$, common choices include absolute residuals $s_j(x,y_j):=|y_j-\hat\mu_j(x)|$ and standardized variants \citep{Diquigiovanni_hyperrectangle,fan2025interpretablemultivariateconformalprediction,hyperectangle_CP}. Another family uses coordinate-wise CQR scores $s_j(x,y_j):=\max\{\hat Q_{j,\alpha/2}(x)-y_j,\, y_j-\hat Q_{j,1-\alpha/2}(x)\}$, where $\hat Q_{j,\alpha/2}$ and $\hat Q_{j,1-\alpha/2}$ are estimated conditional quantiles of $Y_j$ \citep{CQR,MCP_zhou2024conformalized}. Since the maximum can be dominated by the coordinate with the largest scale, \citet{hyperectangle_CP} propose a weighted variant, while \citet{2025maxrank} apply rank transformations before aggregation. This max-score class is the one we aim to cover in our \textit{representation theorem} under some assumptions on the univariate scores  (Section~\ref{sec:rect_implicit}).

\paragraph{Coordinate-wise calibration.}
A related family calibrates each coordinate separately from a multiple-testing perspective \citep{vovk2013transductive}. One computes, for each $j\in[d]$, the conformal quantile $\hat q_{\alpha_j}$ of $\{s_j(X_i,Y_{ij})\}_{i=1}^n$ with $\sum_{j=1}^d\alpha_j\le\alpha$, and returns $\bigtimes_{j=1}^d\{y_j: s_j(x,y_j)\le \hat q_{\alpha_j}\}$. A union bound (Bonferroni correction) guarantees joint coverage at least $1-\alpha$, typically at the cost of conservativeness \citep{vovk2013transductive,dcp_mt,messoudi_NN_bonf}.

\paragraph{Flexible multi-output methods.}
Beyond hyperrectangles, many multi-output CP methods construct flexible sets by modeling dependence across outputs, including copula-based \citep{messoudi2021copula}, density-based \citep{sadinle2019least,izbicki2022cd,wang2022probabilistic}, ellipsoidal and template-based \citep{messoudi2022ellipsoidal,braun2025minimumvolumeconformalsets,braun2025multivariate,tumu2024multi}, and flow- or optimal-transport-based constructions \citep{thurin2025optimal,colombo2024flow,fang2025contra}; see \citet{Dheur2025-unified} for a review. A key principle behind these methods is that the conditional density encodes output dependence: the minimum-volume region with conditional coverage $1-\alpha$ is the density superlevel set $\mathcal{C}^{\mathrm{HDR}}_\alpha(x):=\{y\in\mathbb{R}^d: p_{Y|X}(y\mid x)\ge t_\alpha(x)\}$ for a suitable threshold $t_\alpha(x)$ \citep{hyndman1996computing,izbicki2022cd}. Similarly, the score $$s(x,y):=-\log p_{Y|X}(y\mid x)$$ is optimal in expected volume \citep{sadinle2019least}. \\

Motivated by this density-based view, the next section connects the simple geometry of GH sets to score and density level sets.

\section{Our Rectangularization Framework}
\label{sec:rectangularization}

We now introduce our rectangularization framework, which turns multivariate
nonconformity score into a score whose conformal prediction sets are GH, while
preserving finite-sample coverage. The key idea is to decouple a multivariate
function into univariate, coordinate-wise maps. We first develop the construction
at the level of scores (Section~\ref{sec:score_rectangularization}) and then
specialize it to density-based scores (Section~\ref{sec:density_rect}).
All proofs are deferred to Appendix~\ref{app:proofs}.

\subsection{Score Rectangularization}
\label{sec:score_rectangularization}

Since CP sets are sublevel sets of the score,
$\mathcal{C}_\alpha(x;s)=\{y:s(x,y)\le \hat q_\alpha\}$,
we study rectangularization through the geometry of sublevel sets.
We first characterize the functions whose sublevel sets are GH.

\begin{proposition}
\label{prop:rectangle_levelsets}
Let $f:\mathbb{R}^d\to\mathbb{R}$ be continuous and bounded below. The sublevel set
$\{y\in\mathbb{R}^d:f(y)\le t\}$ is a GH for every $t\in\mathbb{R}$ if and only if
there exist functions $f_1,\dots,f_d:\mathbb{R}\to\mathbb{R}$ such that
\[
f(y)=\max_{j\in[d]} f_j(y_j) \qquad \forall\, y\in\mathbb{R}^d .
\]
\end{proposition}

Thus, functions with GH sublevel sets are exactly those that decouple into a
maximum of univariate functions. This motivates the following construction,
which builds from any NCS $s$ a new score $s^H$ whose sublevel sets are GH.

\begin{definition}[Score rectangularization]
\label{def:rhs}
Let $s:\mathcal{X}\times\mathbb{R}^d\to\mathbb{R}$ be a continuous NCS such that $s(x,\cdot)$
is bounded below for every $x\in\mathcal{X}$. For $j\in[d]$, define the
coordinate scores
\begin{equation}
s_j^H(x,y_j):=\inf_{y_{-j}\in\mathbb{R}^{d-1}} s(x,y),
\end{equation}
The rectangularized score is
\begin{equation}
s^H(x,y):=\max_{j\in[d]} s_j^H(x,y_j).
\end{equation}
\end{definition}

By construction, the CP set of a rectangularized score is always a
GH:
\begin{equation}
\label{eq:GH_predset_s^H}
\begin{aligned}
\mathcal{H}_\alpha(x;s^H)
&:=\left\{y\in\mathbb{R}^d: s^H(x,y)\le \hat q_\alpha^H\right\}\\
&=\bigtimes_{j=1}^d \left\{y_j\in\mathbb{R}: s_j^H(x,y_j)\le \hat q_\alpha^H\right\},
\end{aligned}
\end{equation}
where $\hat q_\alpha^H$ is the conformal quantile of the calibration scores
$\{s^H(X_i,Y_i)\}_{i=1}^n$.

The construction of $s^H$ has a direct geometric interpretation, which yields
the following optimality property.

\begin{proposition}
\label{prop:strict_product_identity}
Fix $x\in\mathcal{X}$ and $t\in\mathbb{R}$. The set
$\{y:s^H(x,y)<t\}=\bigtimes_{j=1}^d\{y_j\in\mathbb{R}:s_j^H(x,y_j)<t\}$
is the smallest GH, with respect to inclusion, containing
$\{y\in\mathbb{R}^d:s(x,y)<t\}$.
Moreover, if $y\mapsto s(x,y)$ is continuous and $\{y:s(x,y)\le t\}$ is compact
for every $t\in\mathbb{R}$, then
\[
\mathcal{H}_\alpha(x;s^H)
\subseteq \bigtimes_{j=1}^d \pi_j\big(\mathcal{C}_\alpha(x;s)\big)
\subseteq \operatorname{RH}\big(\mathcal{C}_\alpha(x;s)\big),
\]
where $\pi_j$ denotes the projection onto the $j$-th coordinate and
$\operatorname{RH}(S)$ is the rectangular hull of $S\subseteq\mathbb{R}^d$,
i.e., the smallest hyperrectangle containing $S$.
\end{proposition}
\begin{table}[t]
\centering
\setlength{\tabcolsep}{4pt}
\renewcommand{\arraystretch}{1.1}
\begin{tabular}{@{}ccc@{}}
\toprule
Score & $s(x,y)$ & $s_{j}^H(x,y_j)$ \\
\midrule
$\ell_p \;(p\in[1,\infty])$
& $\|y - \mu(x)\|_p$
& $|y_j - \mu_j(x)|$ \\
Mahalanobis
& $\|\Sigma(x)^{-1/2}(y - \mu(x))\|_2$
& $\frac{|y_j - \mu_j(x)|}{\sqrt{\Sigma_{jj}(x)}}$ \\
PCP
& $\min\limits_{l\in[L]} \|y - \tilde Y^{(l)}\|_2$
& $\min\limits_{l\in[L]} |y_j - \tilde Y_j^{(l)}|$ \\
Density
& $-\log \mathcal{N}(y; \mu(x), \Sigma(x))$
& $\frac{(y_j - \mu_j(x))^2}{2\Sigma_{jj}(x)} + c(x)$\\
\bottomrule
\end{tabular}
\caption{Rectangularization of common NCSs. Here $\mu$ and $\Sigma$ denote
(estimated) conditional mean and covariance, $\tilde Y^{(l)}$, $l\in[L]$, are
samples from the (estimated) conditional distribution of $Y$ given $X=x$, and
$c(x):=\frac{1}{2}\log\det(2\pi\Sigma(x))$.}
\label{tab:s^H_scores}
\end{table}

%The first statement shows that rectangularization is the tightest GH relaxationof each (strict) sublevel set of $s$. The second follows because$s^H\le s$ pointwise, so that $\hat q_\alpha^H\le\hat q_\alpha$.
%Note that $\operatorname{RH}(\mathcal{C}_\alpha(x;s))$ is itself a valid, butconservative, hyperrectangular prediction set. 
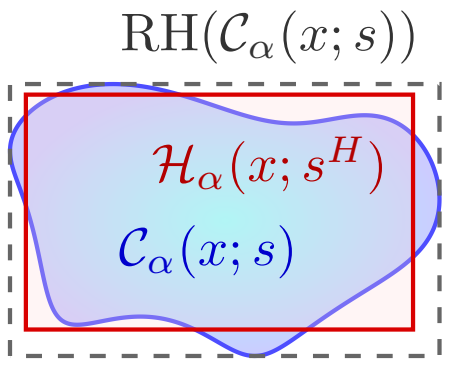
\begin{figure}[ht]
    \centering
    \begin{minipage}{0.58\textwidth}
    Therefore, we deduce from Proposition ~\ref{prop:strict_product_identity} that our construction 
\emph{combines rectangularization with conformal recalibration}: rather than
taking the rectangular hull of $\mathcal{C}_\alpha(x;s)$, we calibrate the
rectangularized score directly. The resulting set is contained in that hull,
while retaining finite-sample coverage. It need not contain
$\mathcal{C}_\alpha(x;s)$ itself (Figure~\ref{fig:rect-levelset}).\end{minipage}
    \hfill
    \begin{minipage}{0.38\textwidth}
\centering
 \begin{tikzpicture}[scale=0.7]

    \begin{scope}[local bounding box=lake]
        \filldraw[
            inner color=cyan!40,
            outer color=blue!20,
            draw=blue!70,
            thick,
            samples=200,
            smooth,
            variable=\t,
            domain=0:360
        ]
        plot ({ (1.6 + 0.4*sin(\t) + 0.2*cos(3*\t)) * cos(\t) * 1.2 },
              { (1.4 + 0.3*cos(2*\t) + 0.2*sin(4*\t)) * sin(\t) }) -- cycle;
    \end{scope}

    % Outer rectangular hull (dashed)
    \draw[thick, dashed, black!60] (lake.south west) rectangle (lake.north east);

    % Rectangularized prediction set H_alpha(x;s^H)
    \filldraw[
        thick,
        draw=red!85!black,
        fill=red!15,
        fill opacity=0.25,
        text opacity=1
    ] ([xshift=0.15cm, yshift=0.25cm] lake.south west) rectangle ([xshift=-0.25cm, yshift=-0.1cm] lake.north east);

    % Labels
    \node[blue!80!black, font=\small] at (0,-0.3) {$\mathcal{C}_\alpha(x;s)$};
    \node[anchor=south east, black!80, font=\small] at (lake.north east) {$\operatorname{RH}(\mathcal{C}_\alpha(x;s))$};
    \node[anchor=north east, red!70!black, font=\small] at ([xshift=-0.3cm, yshift=-0.3cm] lake.north east) {$\mathcal{H}_\alpha(x;s^H)$};

\end{tikzpicture}
 \caption{Illustration of Proposition~\ref{prop:strict_product_identity}.}
   \label{fig:rect-levelset}

    \end{minipage}
\end{figure}

Score rectangularization thus provides a generic way to obtain GH prediction
sets from any multivariate NCS. Table~\ref{tab:s^H_scores} lists common
scores~\citep{ellipse1,messoudi2022ellipsoidal,wang2022probabilistic} and their
rectangularizations. The resulting sets are
interpretable, since each coordinate set shows where the corresponding output
may fall (Figure~\ref{fig:GH_example}, and Figure~\ref{fig:1D_2D_taxi} in
Section~\ref{sec:experiments}). We summarize the steps to compute a GH prediction set from a multivariate NCS\footnote{When $s_i^H$ has no closed form, each evaluation requires a
$(d-1)$-dimensional minimization. The coordinate sets in the last step are
one-dimensional and can be computed on a grid.} in Algorithm~\ref{alg:SCP_s^H}.

\begin{algorithm}[ht]
\caption{SCP with a rectangularized NCS}\label{alg:SCP_s^H}
\begin{algorithmic}[1]
\Require NCS $s$, calibration data $\{(X_i,Y_i)\}_{i=1}^{n}$, test input $X_{n+1}$, $\alpha\in(0,1)$
\For{$j\in [d]$}
    \State Define $s^H_j(x,y_j) \gets \inf_{y_{-j}} s\big(x,(y_j,y_{-j})\big)$
\EndFor
\State Define $s^H(x,y) \gets \max_{j\in[d]} s^H_j(x,y_j)$
\State $\hat{q}_\alpha^H \gets$ $\lceil (n+1)(1-\alpha)\rceil$-th smallest value of $\{s^H(X_i,Y_i)\}_{i=1}^n$
\State \Return $\bigtimes_{j=1}^d \left\{ y_j \in \mathbb{R} : s_j^H(X_{n+1},y_j) \le \hat{q}^H_\alpha\right\}$
\end{algorithmic}
\end{algorithm}

\paragraph{Finite-sample coverage.}
Since $s^H$ is a fixed function of $(x,y)$, the usual split-conformal
exchangeability argument applies directly to the rectangularized scores.

\begin{proposition}
\label{prop:p^H_r^H_coverage}
Assume that $\mathcal{D}_{\mathrm{cal}} \cup \{(X_{n+1}, Y_{n+1})\}$ is
exchangeable. For any NCS $s$ satisfying the conditions of
Definition~\ref{def:rhs}, the prediction set $\mathcal{H}_\alpha(X_{n+1}; s^H)$
is a GH and satisfies
\[
\mathbb{P}\big(Y_{n+1}\in\mathcal{H}_\alpha(X_{n+1};s^H)\big)\ge 1-\alpha .
\]
\end{proposition}

\paragraph{Conditional coverage.}
Our rectangularization can be viewed as a coordinate-wise projection of the
sublevel sets of the original score. This raises the question of how it affects
conditional coverage after recalibration, since the geometric relation between
the two sets does not by itself imply similar conditional coverage. We
nevertheless obtain an explicit comparison under
Gaussianity\footnote{Finite-sample conditional coverage is impossible without
distributional assumptions~\citep{lei2018distribution}.}.

\begin{proposition}
\label{prop:gaussian_conditional_coverage}
Let $\alpha\in(0,1)$ and suppose that
$Y\mid X=x\sim\mathcal N_d(\mu(x),\Sigma(x))$ with $\Sigma(x)\succ0$ for every
$x$. Let $s_{\mathrm{Maha}}$ be the Mahalanobis score based on $\mu(x)$ and
$\Sigma(x)$. Then, for every $x$,
%\[(1-\alpha)^d\le\mathbb P\big(Y\in\mathcal H_\alpha^\star(x;s_{\mathrm{M}}^H)\mid X=x\big)\le(1-\alpha)^{1/d}.\]Consequently, since $\mathcal C_\alpha^\star(x;s_{\mathrm{M}})$ has conditionalcoverage exactly $1-\alpha$,
\[
\begin{aligned}
&\Big|\mathbb P\big(Y\in\mathcal H_\alpha^\star(x;s_{\mathrm{Maha}}^H)\mid X=x\big)
-\mathbb P\big(Y\in\mathcal C_\alpha^\star(x;s_{\mathrm{Maha}})\mid X=x\big)\Big|\\
&\qquad\le \max\left\{(1-\alpha)-(1-\alpha)^d,\,
(1-\alpha)^{1/d}-(1-\alpha)\right\}.
\end{aligned}
\]
\end{proposition}

The bound holds uniformly over $x$ and over all covariance structures, and
depends only on $\alpha$ and $d$. %For instance, with $d=2$ and $\alpha=0.1$, theconditional coverage of the rectangularized set differs from $1-\alpha$ by atmost $0.09$. The bound becomes less informative as $d$ grows. 
It illustrates the intuition that rectangularization can
retain part of the original score's adaptivity to the feature vector, although this is not a universal rule.\\ 

Table~\ref{tab:s^H_scores} also shows that rectangularization discards
structural information, as expected. For example, all $\ell_p$-norm scores of
the residuals induce the same rectangularized score, and hence the same
prediction set. Similarly, the rectangularized Mahalanobis score depends on
$\Sigma(x)$ only through its diagonal entries, ignoring dependence across
outputs. The rectangularized Gaussian density score depends on it only through
its diagonal and its determinant. Consequently, this supports the intuition that rectangularized NCS increases the prediction set volume.

\begin{figure}
    \centering
    \includegraphics[width=0.6\linewidth]{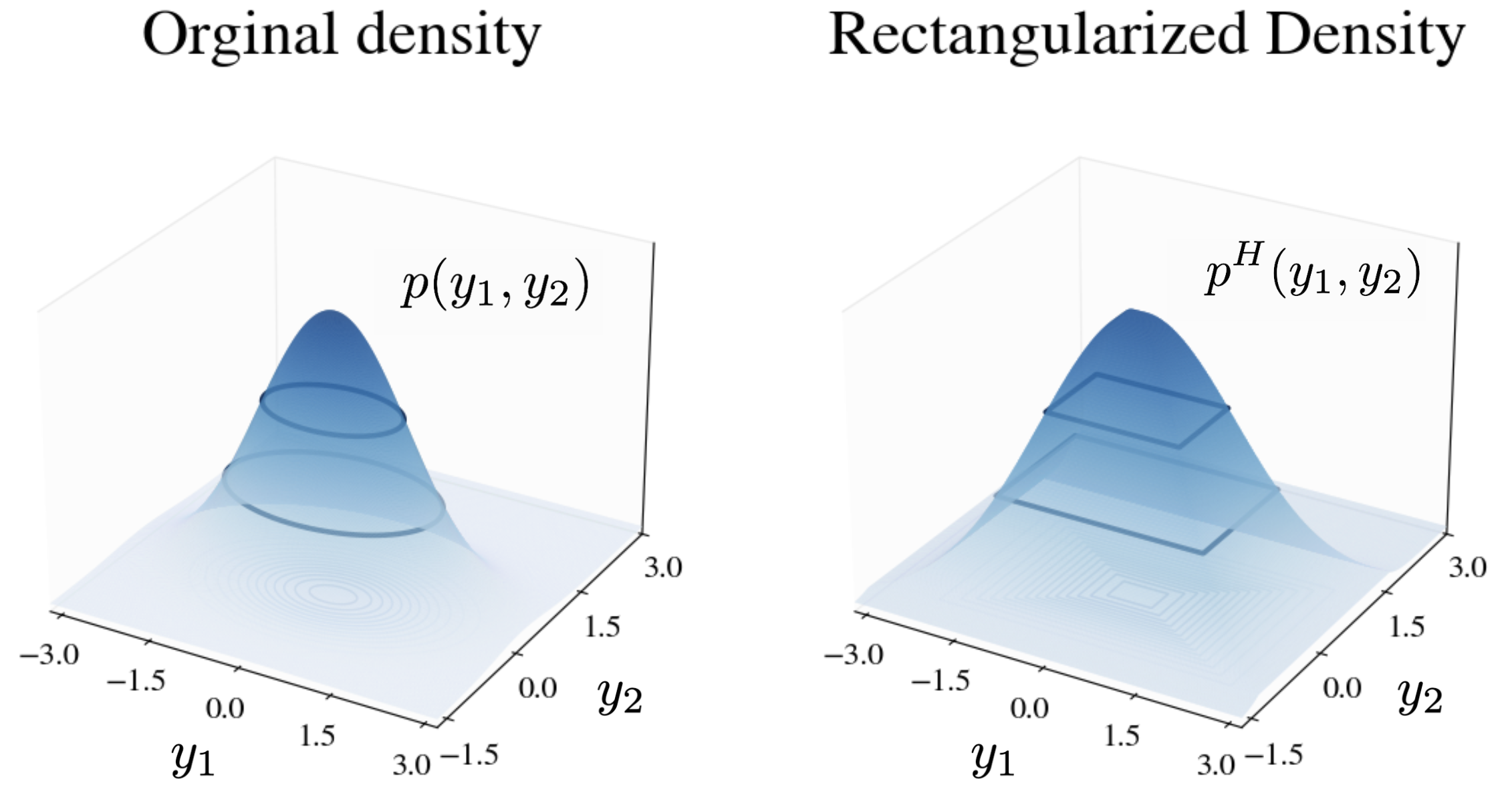}
    \caption{A bivariate Gaussian density $p$ and its rectangularization $p^H$,
    whose superlevel sets are rectangles.}
    \label{fig:3D_illustration_RE}
\end{figure}

\subsection{Density Rectangularization}
\label{sec:density_rect}

Several multi-output CP methods construct nonconformity scores from a
conditional density $p(y\mid x)$, or an estimate thereof. One possible choice is
the negative log-density
$s_p(x,y):=-\log p(y\mid x)$~\citep{sadinle2019least,Dheur2025-unified}. Assuming $p(\cdot\mid x)$ is bounded for every
$x$, applying Definition~\ref{def:rhs} gives
\[
\begin{aligned}
s_p^H(x,y)
&=\max_{j\in[d]}\inf_{y_{-j}}\bigl[-\log p(y\mid x)\bigr]\\
&=-\log\Big(\min_{j\in[d]}\sup_{y_{-j}}p(y\mid x)\Big),
\end{aligned}
\]
Inspired by this identity, we introduce the \emph{rectangularized density} of $p$, denoted by $p^H$ and defined as:
\[
p^H(y\mid x):=\min_{j\in[d]}p_j^H(y_j\mid x),
\qquad
p_j^H(y_j\mid x):=\sup_{y_{-j}}p(y\mid x),
\]
so that $s_p^H=-\log p^H$. The superlevel sets of $p^H$ are GH:
\[
\{y:p^H(y\mid x)\ge t\}
=
\bigtimes_{j=1}^d
\{y_j:p_j^H(y_j\mid x)\ge t\}.
\]
Consequently, conformal calibration of $s_p^H$ yields the GH prediction set
\begin{equation}
\label{eq:H_alpha(-logp^H)}
\mathcal H_\alpha(x;s_p^H)
=
\bigtimes_{j=1}^d
\left\{
y_j:p_j^H(y_j\mid x)\ge e^{-\hat q_\alpha^H}
\right\},
\end{equation}
where $\hat q_\alpha^H$ is computed from the calibration scores
$\{s_p^H(X_i,Y_i)\}_{i=1}^n$. Density rectangularization thus connects
density-based multivariate CP with coordinate-wise prediction sets, while
retaining finite-sample marginal coverage
(Proposition~\ref{prop:p^H_r^H_coverage}).

\paragraph{Example.}
If $p(\cdot\mid x)=\mathcal N_d(\cdot;\mu(x),\Sigma(x))$ with $\Sigma(x)\succ0$, then
$p_j^H(y_j\mid x)
\propto
\frac{1}{(2\pi)^{d/2}\sqrt{\det\Sigma(x)}}
\exp\!\Big(-\frac{(y_j-\mu_j(x))^2}{2\Sigma_{jj}(x)}\Big).$
Each $p_j^H$ is therefore proportional to the univariate Gaussian marginal of
$Y_i$. Figure~\ref{fig:3D_illustration_RE} shows a bivariate Gaussian density and
its rectangularization, whose superlevel sets are rectangles\footnote{One can also show that density rectangularization inherits the minimality property of
Proposition~\ref{prop:strict_product_identity}, where we can provide analogous results.}.

\begin{remark}
Note that $p^H(\cdot\mid x)$ is not a probability density in general. This is harmless for CP, since $p^H$ is only used through
the score $s_p^H=-\log p^H$.
\end{remark}

\section{A Representation Theorem for Max-Aggregated Scores}
\label{sec:rect_implicit}
Having studied the rectangularization of nonconformity scores and conditional densities, we now turn to the converse direction. Our goal is to connect, within our framework, the density-based view of CP with hyperrectangular conformal prediction methods, two approaches that may at first appear quite different. This leads to the following question:
 
\begin{center}
    
\textit{Given a CP method that yields GH prediction sets, does there exist a conditional density whose rectangularization recovers it?}\end{center}
 
\noindent In this section, we focus on CP methods based on the max-aggregated score

$$ s_{\max}(x,y):=\max_{j\in[d]} s_j(x,y_j), $$

where \(s_j:\mathcal{X}\times\mathbb{R}\to\mathbb{R}\) are univariate NCS.
Under suitable assumptions, we establish a \textit{representation theorem}, showing that such methods can indeed be represented through the rectangularization of an implicit conditional density.
 
\paragraph{Assumptions (S).}
For each $j\in[d]$, the score $s_j:\mathcal X\times\mathbb R\to[0,\infty)$ is
measurable, nonnegative, and satisfies:\\[4pt]
(A1) there exists a measurable $m_j:\mathcal X\to\mathbb R$ such that
$s_j(x,m_j(x))=0$ for all $x\in\mathcal X$;\\[4pt]
(A2) there exists $L_j<\infty$ such that
$|s_j(x,z)-s_j(x,z')|\le L_j|z-z'|$ for all $x\in\mathcal X$ and
$z,z'\in\mathbb R$.\\
 
Assumption (A1) states that perfect conformity is achieved at a reference value
$m_j(x)$. Assumption (A2) states that the score varies in a controlled way with
the output, uniformly in $x$. 
Assumptions~(S) hold for common NCSs of the literature (absolute and standardized residuals, CQR score under mild modifications and assumptions), see~\citet{review_cp_scores}
for a review of some univariate scores.
 
\begin{theorem}
\label{main_represetation_theorem}
For $d \geqslant 2$, let $s_1,\dots,s_d$ be nonconformity scores satisfying Assumption \textbf{(S)}, and define for any $y = (y_1,\dots,y_d) \in \mathbb{R}^d$ the max-aggregated multivariate score $s_{\max}(x,y) := \max\{s_1(x,y_1),\dots,s_d(x,y_d)\}$. Then there exists a conditional pdf $\tilde{p}$ such that:
\[
\mathcal{H}_\alpha(x; s_{\max}) = \mathcal{H}_\alpha(x; -\log \tilde{p}^H), \quad \forall x \in \mathcal{X}.
\]
\end{theorem}
 
Consequently, Theorem~\ref{main_represetation_theorem} gives a density-based reading of
CP max-score methods. Through the lens of the theorem, we can
view choosing the univariate NCSs $s_1,\dots,s_d$ as working with a hypothetical conditional density $\tilde p$ whose
rectangularization reproduces the original max-score
construction exactly. This connects two ways of constructing prediction sets from the same underlying density:
 using $-\log\tilde p$ as the score gives a density superlevel
set, typically with complex geometry, whereas using its rectangularization
$-\log\tilde p^H$ gives exactly the GH set $\mathcal H_\alpha(x;s_{\max})$. We
note that the representing density need not be unique,
nor is it assumed to coincide with the true conditional
density of the data. Its role is to provide a common mathematical
interpretation linking coordinate-wise conformal methods to density-based
multivariate CP. 

While we do not claim that this holds for any scores or
any hyperrectangular conformal method, our theorem
constitutes a first step towards unifying different constructions and geometries in CP through this density-based view. The following example illustrates our representation Theorem \ref{main_represetation_theorem} and makes the representation explicit.
 
\begin{proposition}
\label{prop:std_as_RH_gaussian}
Let
$s_{\mathrm{std}}(x,y)=\max_{j\in[d]}|y_j-\mu_j(x)|/\sigma_j(x)$,
where $\mu$ and $\sigma_1,\dots,\sigma_d$ are measurable, $\sigma_j(x)>0$, and
$c:=\inf_{x\in\mathcal X}\prod_{j=1}^d\sigma_j^2(x)>0$. Then there exists $x\mapsto\Sigma(x)\succ0$ with
$\Sigma_{jj}(x)=\sigma_j^2(x)$ for all $j\in[d]$ such that
\[
\mathcal{H}_\alpha(x;s_{\mathrm{std}})
=
\mathcal{H}_\alpha(x;-\log\phi^H)
\qquad\forall\,x\in\mathcal X,
\]
where $\phi(\cdot\mid x):=\mathcal{N}(\cdot\,;\mu(x),\Sigma(x))$.
\end{proposition}
 
Thus, Proposition \ref{prop:std_as_RH_gaussian} provides an interpretation of the $s_{\text{std}}$ CP method as a representation from a multivariate Gaussian model, where the implicit density is $\phi$.
The absolute residual NCS
$s_{\mathrm{abs}}(x,y)=\max_{j\in[d]}|y_j-\mu_j(x)|$ is the special case
$\sigma_j\equiv\sigma$.
Any isotropic Gaussian model $\psi(\cdot\mid x):=\mathcal N(\cdot\,;\mu(x),\sigma^2 I_d)$,
$\sigma>0$, represents it:
\[
\mathcal{H}_\alpha(x;s_{\mathrm{abs}})
=
\mathcal{H}_\alpha(x;-\log\psi^{H})
\qquad\forall\,x\in\mathcal X.
\]

\section{Experiments}
\label{sec:experiments}

We conduct numerical experiments to quantify the effect of rectangularization,
comparing multivariate NCSs with their rectangularized counterparts.

\paragraph{Datasets.}
We consider two synthetic data-generating processes (DGPs) and the real-world
multi-output regression benchmarks of~\citet{Dheur2025-unified}\footnote{code available at \url{https://github.com/Nabil-Ala/Rect_CP_score}.}. Each dataset is
split into $60\%$ training, $20\%$ calibration, and $20\%$ test data. For the
synthetic DGPs, we generate $N=5000$ samples with covariates
$X\sim\mathcal{N}_6(0,\Sigma_X)$, where $\Sigma_X$ is a Toeplitz matrix with
correlation parameter $\rho=0.5$, and vary the output dimension
$d\in\{2,4,8,10,14\}$. In DGP~1, $Y\mid X\sim\mathcal{N}_d(f(X),V(X))$, where
$V(X)$ induces covariate-dependent variances and correlations. In DGP~2,
$Y\mid X$ follows a two-component Gaussian mixture, possibly bimodal, with
$X$-dependent weights and mode separation. Full details are given in
Appendix~\ref{app:experiments}, together with results for two additional
DGPs and supplementary comparisons among existing hyperrectangular CP on real-world
datasets (Appendix~\ref{app:experimental_details}).

\paragraph{Base model.}
All methods share a common base model: a conditional Gaussian mixture model
(GMM)~\citep{cgmm2024},
$\hat{p}(y\mid x)=\sum_{r=1}^R \pi_r(x)\,\mathcal{N}\big(y;\mu_r(x),\Sigma_r\big)$,
fitted on the training set, with $R=3$ for all datasets. The GMM
provides the conditional mean and covariance in closed form and allows sampling
from $\hat{p}(\cdot\mid x)$.

\paragraph{Methods.}
We consider three multivariate NCSs: the p-norm of residuals ($\ell_p$) where we fix $p=2$, the
Mahalanobis score ($\mathrm{Maha}$), and PCP~\citep{wang2022probabilistic}. % TODO: specify the value of L.
We compare each with its rectangularization, denoted $\ell_p^H$\footnote{Note that $\ell_2^H$ coincides with the
$\ell_\infty$ score.},
$\mathrm{Maha}^H$, and $\mathrm{PCP}^H$ (Definition~\ref{def:rhs} and
Table~\ref{tab:s^H_scores}).  We set $\alpha=0.1$ throughout.

\paragraph{Metrics.}
We measure the efficiency cost of rectangularization by the median-log difference of
prediction-set volumes on the test set,
\[
\Delta_V=\operatorname{median}\{\log\vol[\mathcal{H}_\alpha(X_i;s^H)]\}_{i=1}^{n_{\mathrm{test}}}-\operatorname{median}\{\log\vol[\mathcal{C}_\alpha(X_i;s)]\}_{i=1}^{n_{\mathrm{test}}},
\]
where $n_{\mathrm{test}}$ is the test-set size.  We assess conditional coverage with
the ERT diagnostic~\citep{braun2025conditional}, using the authors'
implementation; lower values indicate better conditional coverage. All methods
have finite-sample marginal coverage by construction
(Proposition~\ref{prop:p^H_r^H_coverage}), so we report empirical coverage in
Appendix~\ref{app:experiments}. All results are averaged over $20$ random seeds.

\begin{figure*}[t]
    \centering
    \includegraphics[width=1\linewidth]{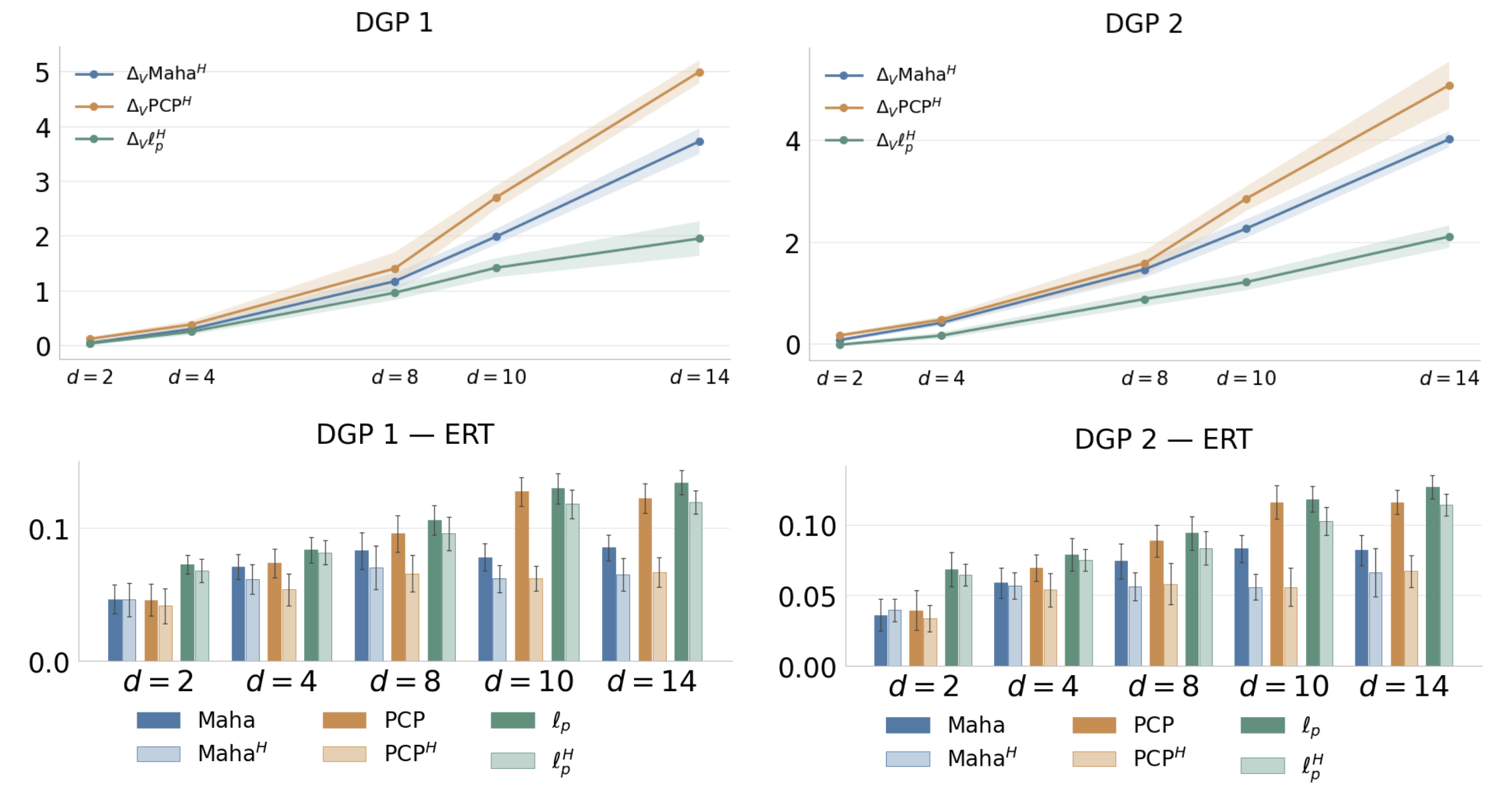}
    \caption{Results on the synthetic DGPs as a function of the output dimension
    $d$, over $20$ random seeds: median log volume difference $\Delta_V$ between each
    rectangularized score and its original score, and ERT.}
    \label{fig:results_simu_s_vs_sH}
\end{figure*}

\paragraph{Results on synthetic data.}
Figure~\ref{fig:results_simu_s_vs_sH} reports the results on the synthetic DGPs.
For low output dimensions, the median volumes of the original and rectangularized
sets are often comparable. As $d$ increases, the gap widens. This is expected,
since the overhead of a product geometry accumulates multiplicatively with $d$,
and hence additively in $\Delta_V$.

\begin{figure}
    \centering
    \includegraphics[width=1\linewidth]{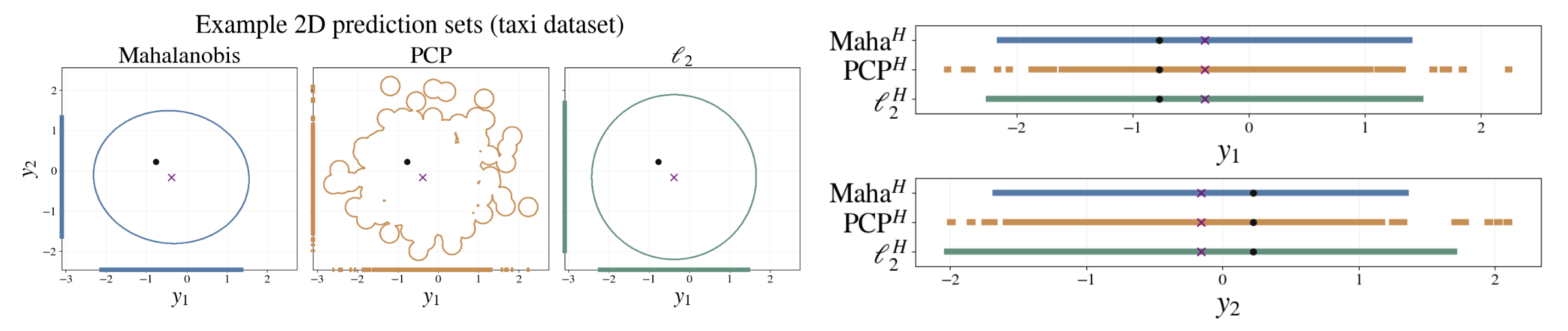}
    \caption{Taxi dataset for one test point. Left: prediction sets of the original scores. Right: coordinate sets of the GH prediction sets induced by the rectangularized scores. Ground truth is shown as black dot.}
    \label{fig:1D_2D_taxi}
\end{figure}

This efficiency cost buys interpretability (Section~\ref{sec:GH_related}). For
$d>3$, a flexible prediction set such as an ellipsoid cannot be visualized
directly, whereas the coordinate sets of a GH can always be displayed separately
in one dimension. Figure~\ref{fig:1D_2D_taxi} illustrates this on the Taxi
dataset, where the goal is to predict a passenger's drop-off location ($d=2$).
Even in this low-dimensional setting, PCP produces cloud-like regions whose
geometry is difficult to read. In contrast, the coordinate sets of the
rectangularized scores show immediately which values are plausible for each
output.

Regarding conditional coverage, rectangularization generally slightly decreases
the ERT. This is consistent with the intuition behind
Proposition~\ref{prop:gaussian_conditional_coverage}, namely that
rectangularization can preserve part of the adaptivity of the original score. It
may also be partly explained by the larger volume of rectangularized sets, which
can sometimes mechanically favor conditional coverage.
\begin{figure}
    \centering
    \includegraphics[width=1\linewidth]{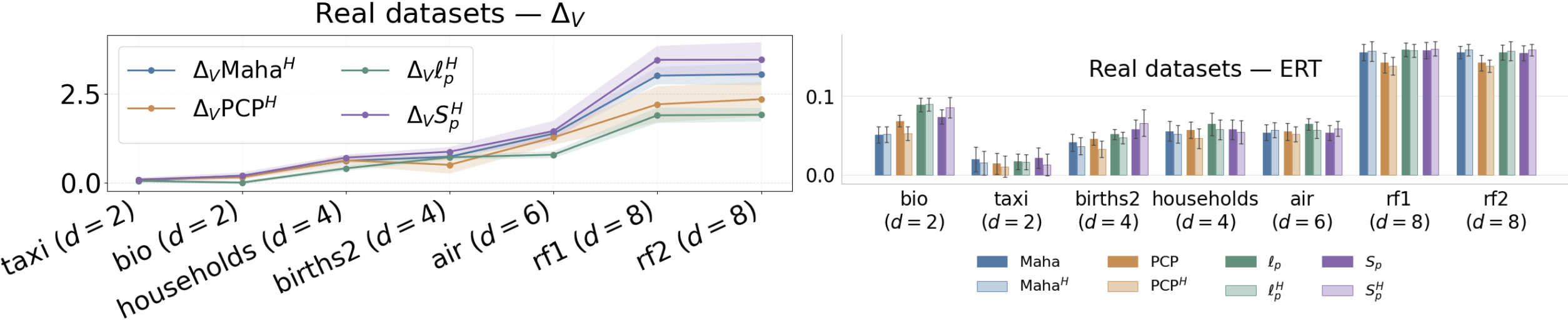}
    \caption{Real-world datasets over $20$ random seeds. Left: 
    $\Delta_V$ plot. Right: ERT of the original and rectangularized scores.}
    \label{fig:s_vs_rect_rw}
\end{figure}

\paragraph{Results on real-world data.}
We evaluate the three nonconformity scores and their rectangularized counterparts on real-world datasets from the literature, where we also add the GMM density score $s(x,y):=-\log \hat p(y|x)$ (denoted in the results as $S_p$) and its rectangularization $s_p^H(x,y):=-\log \hat p^H(y|x)$ ($S_p^H$).

Figure~\ref{fig:s_vs_rect_rw} reports the results where the same pattern emerges: the volume difference grows with the output dimension. On
\texttt{rf2} ($d=8$), $\Delta_V\approx 3$ for the density score, i.e., the median GH set is about
$e^3\approx 20$ times larger than the median original set. Rectangularized sets
again have slightly lower ERT, although the differences are smaller than on
synthetic data.%\footnote{This small conditional coverage difference also goes with the intuition supported byProposition~\ref{prop:gaussian_conditional_coverage}. For example, in the Gaussiansetting with $\alpha=0.1$, Proposition~\ref{prop:gaussian_conditional_coverage} limits the difference in conditional coverage to atmost $9\%$. }.

Overall, these experiments show that rectangularization turns multivariate
scores into GH prediction sets with finite-sample coverage, at a volume cost that
is moderate in low dimensions, but that can become significant as $d$ increases, in exchange for
coordinate-wise interpretability which is very relevant in practical applications of uncertainty quantification.

\section*{Conclusion}
\label{sec:conclusion}

We introduced a rectangularization framework that turns a multivariate nonconformity score, or conditional density, into a score whose conformal prediction sets are generalized hyperrectangles with finite-sample coverage. We established key theoretical properties of our rectangularization framework and provide useful geometric interpretations and insights.  Conversely, our representation theorem shows that, under suitable conditions, max-of-scores hyperrectangular methods arise from the rectangularization of an implicit conditional density, providing a common mathematical interpretation linking coordinate-wise conformal
methods to density-based multivariate CP. Our experiments illustrate the volume cost of rectangularization, and  highlight its benefits, especially for practical applications.

We acknowledge the following limitations. Our representation theorem is only existential and should therefore be viewed primarily as a mathematical tool for unifying multivariate conformal prediction methods. Computing the rectangularized density may not always admit a closed-form expression. While this does not affect the theoretical construction, it can make the procedure more computationally demanding.\\

We believe that hyperrectangular conformal prediction is a rich direction for further research and may contribute to the development of reliable multi-output uncertainty quantification.

\bibliography{references}

\section{Proofs}
\label{app:proofs}
The following lemma will be useful in what follows:
\begin{lemma}
\label{lemma:set_preservation_increasing_function}
Let $g : \mathbb{R} \to \mathbb{R}$ be strictly increasing (independent of $x$) and $s$ a nonconformity score . Then replacing $s$ by $g \circ s$ leaves the split conformal prediction set unchanged, i.e.
\[
C_\alpha(X_{n+1};  s) = C_\alpha(X_{n+1};  g \circ s).
\]
\end{lemma}

\begin{proof}
Denote the calibration scores
\[
S_j := s(X_j, Y_j), \quad j =1,\dots,n,
\]
 and let $\hat{q}_\alpha(s)$ denote the conformal threshold computed from $(S_i)_{i \in \mathcal{I}_{cal}}$, namely the upper order statistic used in split conformal prediction. Since $g$ is strictly increasing, it preserves the ordering of the calibration scores. Note that if $\lceil (n+1)(1-\alpha)\rceil>n$, both prediction sets equal the entire response space. Otherwise, the order statistics of $(g(S_i))_{i \in \mathcal{I}_{cal}}$ are exactly the images under $g$ of the order statistics of $(S_i)_{i \in \mathcal{I}_{cal}}$. In particular,
\[
\hat{q}_\alpha(g \circ s) = g\!\left(\hat{q}_\alpha(s)\right).
\]
Now, by definition,
\[
C_\alpha(X_{n+1};  s) = \left\{y : s(X_{n+1}, y) \leq \hat{q}_\alpha(s)\right\},
\]
while
\[
C_\alpha(X_{n+1};  g \circ s) = \left\{y : g(s(X_{n+1}, y)) \leq \hat{q}_\alpha(g \circ s)\right\}.
\]
Using the identity above, this becomes
\[
C_\alpha(X_{n+1};  g \circ s) = \left\{y : g(s(X_{n+1}, y)) \leq g(\hat{q}_\alpha(s))\right\}.
\]
Since $g$ is strictly increasing, $g(s(X_{n+1}, y)) \leq g(\hat{q}_\alpha(s)) \Longleftrightarrow s(X_{n+1}, y) \leq \hat{q}_\alpha(s)$. Hence
\[
C_\alpha(X_{n+1};  g \circ s) = \left\{y : s(X_{n+1}, y) \leq \hat{q}_\alpha(s)\right\} = C_\alpha(X_{n+1};  s). \qedhere
\]
\end{proof}

\paragraph{Proof of Proposition~\ref{prop:rectangle_levelsets}.}
If $f(y)=\max_i f_i(y_i)$, then
\[
\{y:f(y)\le t\}
=
\bigtimes_{i=1}^d\{y_i:f_i(y_i)\le t\},
\]
which is a GH set.

Conversely, suppose every sublevel set of $f$ is a GH set.
Define
\[
f_i(u):=\inf_{z_{-i}}f(u,z_{-i}),
\qquad
g(y):=\max_{1\le i\le d}f_i(y_i).
\]
Each $f_i$ is finite because $f$ is real-valued and
bounded below. Moreover, $g(y)\le f(y)$.

Fix $y\in\mathbb R^d$ and $\varepsilon>0$, and set
$t=g(y)+\varepsilon$.
For each $i$, the definition of the infimum gives
a vector $z^{(i)}$ satisfying
$z_i^{(i)}=y_i$ and $f(z^{(i)})<t$.
Thus $y_i$ belongs to the $i$th coordinate projection
of $L_t:=\{z:f(z)\le t\}$.
Since $L_t$ is a GH set,
\[
y\in\prod_{i=1}^d\pi_i(L_t)=L_t.
\]
Hence $f(y)\le g(y)+\varepsilon$.
Letting $\varepsilon\downarrow0$ gives $f(y)\le g(y)$,
and therefore $f=g$.
\hfill$\square$

\paragraph{Proof of Proposition \ref{prop:strict_product_identity}}

\begin{proof}
Let $A_t=\{y\in\mathbb R^d:s(x,y)<t\}$ and denote by
$\pi_i(A_t)$ its projection onto coordinate $i$.
By the definition of the infimum,
\[
s_i^H(x,y_i)<t
\quad\Longleftrightarrow\quad
\exists\,y_{-i}\in\mathbb R^{d-1}:s(x,y)<t.
\]
Consequently,
\[
\{y:s^H(x,y)<t\}=\bigtimes_{i=1}^d\pi_i(A_t).
\]
This product contains $A_t$. Moreover, if a generalized
hyperrectangle $G=\bigtimes_{i=1}^d G_i$ contains $A_t$,
then $\pi_i(A_t)\subseteq G_i$ for every $i$. Hence
$\bigtimes_{i=1}^d\pi_i(A_t)\subseteq G$, proving minimality.
We now prove the conformal-set inclusion. Since \(s^H(x,y)\le s(x,y)\) for all \(y\), the calibration scores satisfy

$$ s^H(X_j,Y_j)\le s(X_j,Y_j),\qquad j=1,\ldots,n, $$

and hence, because the split-conformal threshold is an order statistic,

$$ \hat q_\alpha^H\le \hat q_\alpha. $$

Let \(y\in \mathcal  H_\alpha(x;s^H)\). Then, for every \(i\),

$$ s_i^H(x,y_i)\le \hat q_\alpha^H\le \hat q_\alpha \implies \inf_{y_{-i}}s(x,y_i,y_{-i})\le \hat q_\alpha. $$

Under the assumptions that \(y\mapsto s(x,y)\) is continuous and $\{y:s(x,y)\le \hat q_\alpha\} $ is compact, the infimum is attained whenever it is finite. Hence there exists \(z_{-i}\) such that $ s(x,y_i,z_{-i})\le \hat q_\alpha, $ so \(y_i\) belongs to the \(i\)-th coordinate projection of \(C_\alpha(x;s)\). Therefore

$$ y\in \bigtimes_{i=1}^d\pi_i(\mathcal C_\alpha(x;s)) \subseteq \operatorname{RH}(\mathcal  C_\alpha(x;s)). $$

\end{proof}

\paragraph{Proof of Proposition \ref{prop:p^H_r^H_coverage}}
\begin{proof}
    
The proof is straightforward from the usual exchangeability argument \citep{vovk2005algorithmic,shafer2008tutorial} of the nonconformity scores $s^H(X_1,Y_1),$ $\dots$$,s^H(X_n,Y_n),s^H(X_{n+1},Y_{n+1}).$
\end{proof}

\paragraph{Proof of Proposition \ref{prop:gaussian_conditional_coverage}}

\begin{proof}
Write $\tau=1-\alpha$. Since
$s_{\mathrm{Maha}}(X,Y)^2\mid X=x\sim\chi_d^2$,
\[
\mathbb P(s_{\mathrm{Maha}}(X,Y)\le q_{\mathrm{Maha}}
          \mid X=x)=\tau.
\]
Let $u(t)=2\Phi(t)-1$ and
$c_H(x;t)=\mathbb P(s^H(X,Y)\le t\mid X=x)$.
Conditionally on $X=x$, the standardized coordinates
are jointly Gaussian with standard normal marginals.
Thus, by \v{S}id\'ak's inequality and event inclusion,
\[
u(t)^d\le c_H(x;t)\le u(t),
\qquad t\ge0.
\]
The marginal distribution of $s^H(X,Y)$ is continuous,
so $\mathbb E_X[c_H(X;q^H)]=\tau$.
Taking expectations in the preceding inequalities gives
\[
u(q^H)^d\le\tau\le u(q^H),
\qquad\text{hence}\qquad
\tau\le u(q^H)\le\tau^{1/d}.
\]
Consequently, for every $x$,
\[
\tau^d\le c_H(x;q^H)\le\tau^{1/d}.
\]
Subtracting $\tau$ proves the claim.
\end{proof}

\paragraph{Proof of Theorem~\ref{main_represetation_theorem}.}
\begin{proof}
Fix the constants $L_i>0$ from (A2), enlarging them if necessary, and fix
\[
0<C<2\Big(\prod_{i=1}^d L_i\Big)^{-1/d}.
\]
Write $m(x):=(m_1(x),\dots,m_d(x))$ and define
\[
p_0(y\mid x):=\exp\!\Big(-C\sum_{i=1}^d s_i(x,y_i)\Big),
\qquad
R(u):=\sum_{i=1}^d\sum_{j\ne i}u_i^2\,|u_j|,\quad u\in\mathbb R^d .
\]
 
\emph{Step 1: integrability of $e^{-\lambda R}$.}
Clearly $R\ge0$, and $R(u)=0$ if and only if at most one coordinate of $u$ is
nonzero. This set is a finite union of coordinate axes, which is Lebesgue-null
since $d\ge2$. Hence $R>0$ almost everywhere. Moreover, for every $i$,
$R(u)\ge u_i^2\sum_{j\ne i}|u_j|$. Splitting $\mathbb R^d$ into $[-1,1]^d$ and
the sets $A_i:=\{u:|u_i|>1\}$, and using $e^{-\lambda R}\le 1$ on the cube, we
obtain for every $\lambda>0$
\[
\begin{aligned}
\int_{\mathbb R^d}e^{-\lambda R(u)}\,du
&\le 2^d+\sum_{i=1}^d\int_{|u_i|>1}\int_{\mathbb R^{d-1}}
\exp\!\Big(-\lambda u_i^2\sum_{j\ne i}|u_j|\Big)\,du_{-i}\,du_i\\
&= 2^d+\sum_{i=1}^d\int_{|u_i|>1}\Big(\frac{2}{\lambda u_i^2}\Big)^{d-1}du_i
= 2^d+\frac{2d}{2d-3}\Big(\frac{2}{\lambda}\Big)^{d-1}<\infty,
\end{aligned}
\]
where we used $\int_{\mathbb R}e^{-a|v|}\,dv=2/a$ for $a>0$ and $2(d-1)>1$.
 
\emph{Step 2: normalization.}
For $\lambda>0$, let
\[
I_x(\lambda):=\int_{\mathbb R^d}p_0(y\mid x)\,e^{-\lambda R(y-m(x))}\,dy .
\]
Since $0<p_0\le1$, Step~1 gives $I_x(\lambda)<\infty$. For $\lambda\ge\lambda_0>0$
the integrand is dominated by the integrable function $e^{-\lambda_0R(y-m(x))}$,
so by dominated convergence $I_x$ is continuous on $(0,\infty)$, and
$I_x(\lambda)\to0$ as $\lambda\to\infty$ because $R>0$ almost everywhere. Since
$p_0>0$ and $R>0$ almost everywhere, $I_x$ is strictly decreasing. By (A1)--(A2),
$s_i(x,m_i(x)+u_i)\le L_i|u_i|$, so by monotone convergence and the choice of $C$,
\[
\lim_{\lambda\downarrow0}I_x(\lambda)
=\int_{\mathbb R^d}p_0(y\mid x)\,dy
\ge\prod_{i=1}^d\int_{\mathbb R}e^{-CL_i|u_i|}\,du_i
=\prod_{i=1}^d\frac{2}{CL_i}>1,
\]
where the integral may be infinite. Hence there exists a unique $\lambda(x)>0$
with $I_x(\lambda(x))=1$. This choice is measurable, because $x\mapsto I_x(q)$ is
measurable for every $q>0$ by Tonelli's theorem and
$\lambda(x)=\inf\{q\in\mathbb Q_{>0}:I_x(q)\le1\}$.
 
\emph{Step 3: rectangularization.}
Define
\[
\tilde p(y\mid x):=p_0(y\mid x)\,e^{-\lambda(x)R(y-m(x))}.
\]
It is measurable, continuous in $y$ by (A2), and integrates to one. Since
$s_j\ge0$ and $R\ge0$, for each $i$,
\[
\tilde p(y\mid x)\le e^{-C s_i(x,y_i)} .
\]
Equality holds when $y_j=m_j(x)$ for every $j\ne i$: the other scores vanish by
(A1), and $y-m(x)$ then has at most one nonzero coordinate, so
$R(y-m(x))=0$. Therefore $\tilde p_i^H(y_i\mid x)=\sup_{y_{-i}}\tilde p(y\mid x)=e^{-Cs_i(x,y_i)}$,
and
\[
-\log\tilde p^H(y\mid x)
=-\log\min_{i\in[d]}e^{-Cs_i(x,y_i)}
=C\max_{i\in[d]}s_i(x,y_i)
=C\,s_{\max}(x,y).
\]
Since $u\mapsto Cu$ is strictly increasing,
Lemma~\ref{lemma:set_preservation_increasing_function} gives
$\mathcal H_\alpha(x;s_{\max})=\mathcal H_\alpha(x;-\log\tilde p^H)$ for every
calibration set and every $x\in\mathcal X$.
\end{proof}
\paragraph{Proof of Proposition \ref{prop:std_as_RH_gaussian}}
\begin{proof}
Fix $x\in\mathcal X$, and define
\[
D(x):=\mathrm{diag}\bigl(\sigma_1(x),\dots,\sigma_d(x)\bigr).
\]
Set
\[
r(x):=\frac{c}{\prod_{i=1}^d \sigma_i^2(x)}\in(0,1].
\]
Since $d\ge 2$, define the correlation matrix
\[
R(x):=
\begin{pmatrix}
1 & \rho(x) \\
\rho(x) & 1
\end{pmatrix}
\oplus I_{d-2},
\qquad
\rho(x):=\sqrt{1-r(x)}.
\]
Then $R(x)$ is positive definite, its diagonal entries are all equal to $1$, and
\[
|R(x)|=1-\rho(x)^2=r(x).
\]
Now define
\[
\Sigma(x):=D(x)\,R(x)\,D(x).
\]
Since $R(x)$ is positive definite and $D(x)$ is invertible, $\Sigma(x)$ is positive definite. Moreover,
\[
\Sigma_{ii}(x)=\sigma_i^2(x),\qquad i=1,\dots,d,
\]
and
\[
|\Sigma(x)|
=
|D(x)|^2\,|R(x)|
=
\Bigl(\prod_{i=1}^d \sigma_i^2(x)\Bigr)\,r(x)
=
c.
\]

Let
\[
p(y\mid x):=\mathcal N\bigl(y;\mu(x),\Sigma(x)\bigr).
\]
We claim that, for each $i$,
\[
p_{i}^H(y_i\mid x)
=
(2\pi)^{-d/2}|\Sigma(x)|^{-1/2}
\exp\!\left(
-\frac{(y_i-\mu_i(x))^2}{2\,\Sigma_{ii}(x)}
\right).
\]
Indeed, write $\delta:=y-\mu(x)$ and $\Omega(x):=\Sigma(x)^{-1}$. Then
\[
p(y\mid x)
=
(2\pi)^{-d/2}|\Sigma(x)|^{-1/2}
\exp\!\left(-\frac12 \delta^\top \Omega(x)\delta\right).
\]
For fixed $y_i$, the supremum over $y_{-i}$ is obtained by minimizing the quadratic form
$\delta^\top\Omega(x)\delta$ with respect to $\delta_{-i}$. A standard Schur complement computation yields
\[
\inf_{y_{-i}} \delta^\top\Omega(x)\delta
=
\frac{(y_i-\mu_i(x))^2}{\Sigma_{ii}(x)}.
\]
Hence the displayed formula for $p_{i}^H$ follows.

Since $\Sigma_{ii}(x)=\sigma_i^2(x)$, we obtain
\[
p^H(y\mid x)
=
\min_{1\le i\le d} p_{i}^H(y_i\mid x)
=
(2\pi)^{-d/2}c^{-1/2}
\exp\!\left(
-\frac12
\max_{1\le i\le d}
\frac{(y_i-\mu_i(x))^2}{\sigma_i^2(x)}
\right).
\]
Therefore
\[
-\log p^H(y\mid x)
=
\frac d2\log(2\pi)+\frac12\log c
+
\frac12
\max_{1\le i\le d}
\frac{(y_i-\mu_i(x))^2}{\sigma_i^2(x)}.
\]
Since
\[
s_{\mathrm{std}}(x,y)
=
\max_{1\le i\le d}\frac{|y_i-\mu_i(x)|}{\sigma_i(x)},
\]
we conclude that
\[
-\log p^H(y\mid x)
=
\frac d2\log(2\pi)+\frac12\log c+\frac12 s_{\mathrm{std}}(x,y)^2.
\]
Thus $-\log p^H$ is a global strictly increasing transform of $s_{\mathrm{std}}$. By the invariance of split conformal prediction under strictly increasing transformations (Lemma \ref{lemma:set_preservation_increasing_function}) of the score, this yields
\[
\mathcal H_\alpha\!\left(x;\mathcal D_{\mathrm{cal}};s_{\mathrm{std}}\right)
=
\mathcal H_\alpha\!\left(x;\mathcal D_{\mathrm{cal}};-\log \mathcal N^H\right).
\]
\end{proof}

For the $s_{\mathrm{abs}}$ score, the proof is a particular case where $\Sigma=\sigma^2 I_d$ is independent of $X$, so that $\Sigma_{ii}=\sigma^2$ for all $i$ and $|\Sigma|=\sigma^{2d}$. Hence
\[
-\log \psi^H(y\mid x)
=
\frac d2\log(2\pi)+d\log\sigma
+
\frac{1}{2\sigma^2}s_{\mathrm{abs}}(x,y)^2,
\]
which is again a strictly increasing transform of $s_{\mathrm{abs}}$, and Lemma~\ref{lemma:set_preservation_increasing_function} gives the claimed equality of conformal sets.

\paragraph{Score Rectangularization}
\begin{table}[ht]
\centering
\caption{Rectangularization of common NCSs used in multi-output CP. }
\label{tab:profile_functions}
\renewcommand{\arraystretch}{1.5}
\begin{tabular}{|l|c|c|}
\hline
\textbf{Score} & $s(x, y)$ & $s_j^H(y_j)$ \\ \hline
$\ell_\infty$ & $\|y - \mu\|_\infty$ & $|y_j - \mu_j|$ \\ \hline
$\ell_p$ Norm & $\|y - \mu\|_p$ & $|y_j - \mu_j|$ \\ \hline
Mahalanobis & $\sqrt{(y - \mu)^T \Sigma^{-1} (y - \mu)}$ & $\frac{|y_j - \mu_j|}{\sqrt{\Sigma_{jj}}}$ \\ \hline
PCP &
$\displaystyle \min_{l\in[L]}\|y-\tilde Y^{(l)}\|_2,\; \tilde Y^{(l)}\sim \hat F_{Y\mid x}$ &
$\displaystyle \min_{l\in[L]}|y_j-\tilde Y_j^{(l)}|$ \\ \hline
 Density &$-\log \mathcal{N}(y; \mu(x), \Sigma(x))$& $\frac{(y_j - \mu_j(x))^2}{2\Sigma_{jj}(x)} + \frac{1}{2}\log\det(2\pi\Sigma(x)).$\\ \hline
\end{tabular}
\end{table}

\begin{proof}
We fix $x$ and write $\mu=\mu(x)$.

\paragraph{$\ell_\infty$ score.}
For $s(y)=\|y-\mu\|_\infty$, we have
\[
s_i^H(y_i)
=
\inf_{y_{-i}}
\max_i|y_i-\mu_i|
=
|y_i-\mu_i|,
\]
where the infimum is attained by taking $y_j=\mu_j$ for all $j\ne i$.

\paragraph{$\ell_p$ score.}
For $s(y)=\|y-\mu\|_p$ with $1\le p<\infty$,
\[
s_i^H(y_i)
=
\inf_{y_{-i}}
\left(|y_i-\mu_i|^p+\sum_{j\ne i}|y_j-\mu_j|^p\right)^{1/p}
=
|y_i-\mu_i|,
\]
again attained at $y_j=\mu_j$ for $j\ne i$. 

The case $p = \infty$ follows similarly, since
\[
\inf_{y_{-i}}\max_{1\le j\le d}|y_j-\mu_j|
=
|y_i-\mu_i|.
\]

\paragraph{Mahalanobis score.}
Let
\[
s(y)=\sqrt{(y-\mu)^\top\Sigma^{-1}(y-\mu)},
\]
where $\Sigma$ is positive definite. Write $\delta=y-\mu$ and
$\Omega=\Sigma^{-1}$. Partition $\delta=(\delta_i,\delta_{-i})$ and $\Omega$
accordingly. For fixed $\delta_i$, minimizing
$\delta^\top\Omega\delta$ over $\delta_{-i}$ gives
\[
\delta_{-i}^\star
=
-\Omega_{-i,-i}^{-1}\Omega_{-i,i}\delta_i.
\]
Hence
\[
\inf_{\delta_{-i}}\delta^\top\Omega\delta
=
\left(
\Omega_{ii}
-
\Omega_{i,-i}\Omega_{-i,-i}^{-1}\Omega_{-i,i}
\right)\delta_i^2.
\]
By the Schur complement identity,
\[
\Omega_{ii}
-
\Omega_{i,-i}\Omega_{-i,-i}^{-1}\Omega_{-i,i}
=
\frac{1}{\Sigma_{ii}}.
\]
Therefore
\[
s_i^H(y_i)
=
\frac{|y_i-\mu_i|}{\sqrt{\Sigma_{ii}}}.
\]

\paragraph{PCP score.}
Let
\[
s_{\mathrm{PCP}}(y)
=
\min_{\ell\in[L]}\|y-\widetilde Y^{(\ell)}\|_2 .
\]
Then
\[
s_{\mathrm{PCP},i}^H(y_i)
=
\inf_{y_{-i}}\min_{\ell\in[L]}
\left(
|y_i-\widetilde Y_i^{(\ell)}|^2
+
\sum_{j\ne i}|y_j-\widetilde Y_j^{(\ell)}|^2
\right)^{1/2}.
\]
Since the minimum over $\ell$ is finite, the infimum and minimum may be exchanged:
\[
s_{\mathrm{PCP},i}^H(y_i)
=
\min_{\ell\in[L]}
\inf_{y_{-i}}
\left(
|y_i-\widetilde Y_i^{(\ell)}|^2
+
\sum_{j\ne i}|y_j-\widetilde Y_j^{(\ell)}|^2
\right)^{1/2}.
\]
For each $\ell$, the inner infimum is attained at
$y_j=\widetilde Y_j^{(\ell)}$ for all $j\ne i$, so
\[
s_{\mathrm{PCP},i}^H(y_i)
=
\min_{\ell\in[L]} |y_i-\widetilde Y_i^{(\ell)}|.
\]

\paragraph{Density score.} 
$$s(x, y) = -\log \mathcal{N}(y;\mu(x),\Sigma(x))= \frac{1}{2} (y - \mu(x))^T \Sigma(x)^{-1} (y - \mu(x)) + \frac{1}{2} \log\det(2\pi \Sigma(x))$$
so we recognize the squared Mahalanobis score,  leading to 

$$s_i^H(x,y)=\frac{(y_i - \mu_i(x))^2}{2\Sigma_{ii}(x)} + \frac{1}{2}\log\det(2\pi\Sigma(x)).$$
\end{proof}

\begin{comment}
Assume their exist a strictly increasing and continuous function $g$ such that $$\forall x\in \mathcal{X}, \forall y\in \mathbb{R}^d,\;|s(x,y)-g\circ w(x,y)|\le \delta(x).$$
Thenfor two multi-output nonconformity scores. Then for all $$-\delta + g(w(x,y))\le s(x,y) \le\delta + g(w(x,y))$$
Because $g$ is strictly increasing and continuous it preserves the infimum, thus
$$\forall i\in[d], \;-\delta + g(w_i^H(x,y))\le s_i^H(x,y) \le\delta + g(w_i^H(x,y))$$
and we can take the max as $g$ also preserves the maximum so we get
$$ -\delta + g(w^H(x,y))\le s^H(x,y) \le\delta + g(w^H(x,y)).$$

By computing the above inequality for the calibration set, the CP quantiles also follow the same inequality:

$$ -\delta(x) + g(\hat q^w_\alpha)\le \hat q^s_\alpha \le\delta(x) + g(\hat q^w_\alpha).$$
so the prediction set difference can be bounded as
$$\mathcal{H}_\alpha(x;s) \Delta \mathcal{H}_\alpha(x;w) \subseteq \left\{y\in\mathbb{R}^d : \big\vert{} s^H(x,y) - \hat{q}^s_\alpha \big\vert{} \le 2\delta(x) \right\}$$
which intuitively is small. 

Thus, 

$$\vol (\mathcal{H}_\alpha(x;s) \Delta \mathcal{H}_\alpha(x;w))\le \vol \mathcal H$$
\end{comment}

\section{ Experiment Details}
\label{app:experiments}

\subsection{Synthetic Data Generation}
\label{app:synthetic_data}

We consider four synthetic multivariate regression settings with
covariates $X\in\mathbb{R}^{6}$ and responses $Y\in\mathbb{R}^{d}$,
where $d\in\{2,4,8,10,14\}$. For each setting and response dimension,
we generate $n=5000$ observations. The four settings differ in their
conditional response distributions: Gaussian noise with fixed
covariance, Student-$t$ noise with fixed scale, Gaussian noise with
covariate-dependent covariance, and a covariate-dependent Gaussian
mixture. We generate these datasets once before evaluating the methods.

\subsection{Common construction}

For each response dimension, the covariates are sampled independently as
\begin{equation}
    X_i \sim \mathcal{N}_6(0,\Sigma_X),
    \qquad
    (\Sigma_X)_{ab}=0.5^{|a-b|},
    \quad a,b=1,\ldots,6.
\end{equation}
The same covariate sample is reused across the four settings at a given
$d$. We define the nonlinear feature map
\begin{equation}
    \phi(x)=
    \bigl(x_1,x_2,\tanh(x_3),x_4^2-1,x_5x_6\bigr)^\top
    \in\mathbb{R}^{5},
\end{equation}
and the common location function
\begin{equation}
    f_d(x)=1.5A_d\phi(x).
\end{equation}

To construct the parameters, we draw master matrices
$A^\star\in\mathbb{R}^{14\times5}$ and
$B_0^\star,B_1^\star,B_2^\star\in\mathbb{R}^{14\times3}$
with independent standard normal entries. Each row $g$ is then
normalized as $g/(\|g\|_2+10^{-12})$. These matrices are drawn once
and held fixed across all datasets. For response dimension $d$, we use
\begin{equation}
    A_d=A^\star_{1:d,\,1:5},
    \qquad
    B_{\ell,d}=(B_\ell^\star)_{1:d,\,1:r_d},
    \qquad
    r_d=\min(3,d),
\end{equation}
for $\ell\in\{0,1,2\}$. No additional row normalization is performed
after truncation.

All observations are generated independently conditional on these
fixed parameters. We write $I_d$ for the identity matrix and
$\varepsilon=10^{-6}$ for the diagonal regularization used in
Cholesky sampling.

\subsection{DGP 1: Gaussian responses with covariate-dependent covariance}

We construct a covariate-dependent covariance through a
low-rank factor contribution and a diagonal contribution:
\begin{align}
    L_d(x)
    &=0.25B_{0,d}+0.18x_1B_{1,d}+0.18x_2B_{2,d},\\
    a(x)
    &=0.20\bigl(1+0.50x_3^2+0.30x_4^2\bigr),\\
    V_d(x)
    &=L_d(x)L_d(x)^\top+\bigl(a(x)+\varepsilon\bigr)I_d.
\end{align}
For each observation, let $C_d(X_i)$ be the lower Cholesky factor
of $V_d(X_i)$, draw $Z_i\sim\mathcal{N}_d(0,I_d)$ independently of
$X_i$, and generate
\begin{equation}
    Y_i=f_d(X_i)+C_d(X_i)Z_i.
\end{equation}
Hence,
\begin{equation}
    Y_i\mid X_i=x
    \sim\mathcal{N}_d\bigl(f_d(x),V_d(x)\bigr).
\end{equation}
Both the conditional variances and the correlations between response
coordinates can vary with the covariates. The diagonal contribution
ensures that $V_d(x)$ is positive definite for every $x$.

\subsection{DGP 2: Covariate-dependent Gaussian mixture}

This setting uses the same component covariance $V_d(x)$ as DGP 1. %DGP1 =case3
Define the mixing probability, displacement multiplier, and
alternating displacement vector by
\begin{align}
    \pi(x)
    &=\frac{1}{1+\exp\{-(x_1-x_2+0.5x_3)\}},\\
    h(x)
    &=\min\bigl\{1.40,\max\{0.25,0.70+0.25x_2\}\bigr\},\\
    (\delta_d)_j
    &=0.60(-1)^{j-1},
    \qquad j=1,\ldots,d.
\end{align}
For each observation, draw
\begin{equation}
    K_i\mid X_i
    \sim\operatorname{Bernoulli}\bigl(\pi(X_i)\bigr),
    \qquad
    Z_i\sim\mathcal{N}_d(0,I_d),
\end{equation}
with $Z_i$ independent of $(X_i,K_i)$, and set
\begin{equation}
    Y_i
    =f_d(X_i)+(2K_i-1)h(X_i)\delta_d+C_d(X_i)Z_i.
\end{equation}
One indicator $K_i$ selects the component jointly for all response
coordinates. The resulting conditional distribution is
\begin{equation}
\begin{aligned}
    Y_i\mid X_i=x \sim{}&
    \pi(x)\mathcal{N}_d
    \bigl(f_d(x)+h(x)\delta_d,V_d(x)\bigr)\\
    &+(1-\pi(x))\mathcal{N}_d
    \bigl(f_d(x)-h(x)\delta_d,V_d(x)\bigr).
\end{aligned}
\end{equation}
The mixing weights, component separation, and component covariance
all depend on the covariates. In particular, the conditional moments
of the full mixture are
\begin{align}
    \mathbb{E}[Y_i\mid X_i=x]
    &=f_d(x)+(2\pi(x)-1)h(x)\delta_d,\\
    \operatorname{Cov}(Y_i\mid X_i=x)
    &=V_d(x)
      +4\pi(x)(1-\pi(x))h(x)^2\delta_d\delta_d^\top.
\end{align}
Thus, $f_d(x)$ is the midpoint of the two component means and generally
differs from the conditional mean. This construction permits
multimodal conditional distributions, although bimodality is not
guaranteed for every covariate value.

\subsection{DGP 3: Gaussian responses with fixed covariance}

Define the fixed covariance matrix
\begin{equation}
    (\Sigma_{0,d})_{jk}
    =0.45^2\,0.6^{|j-k|},
    \qquad j,k=1,\ldots,d.
\end{equation}
Responses are generated according to
\begin{equation}
    Y_i=f_d(X_i)+E_i,
    \qquad
    E_i\sim\mathcal{N}_d(0,\Sigma_{0,d}),
\end{equation}
where the residuals are independent of the covariates. Equivalently,
\begin{equation}
    Y_i\mid X_i=x
    \sim\mathcal{N}_d\bigl(f_d(x),\Sigma_{0,d}\bigr).
\end{equation}
This setting has a nonlinear conditional mean and a constant
conditional covariance. Each response coordinate has conditional
noise standard deviation $0.45$. No diagonal regularization is added
in this case.

\subsection{DGP 4: Student-\texorpdfstring{$t$}{t} responses with fixed scale}

Let
\begin{equation}
    S_{0,d}=\Sigma_{0,d}+\varepsilon I_d,
    \qquad
    C_{0,d}C_{0,d}^{\top}=S_{0,d},
\end{equation}
where $C_{0,d}$ is the lower Cholesky factor. For each observation,
draw independently
\begin{equation}
    Z_i\sim\mathcal{N}_d(0,I_d),
    \qquad
    U_i\sim\chi^2_5,
\end{equation}
independently of $X_i$, and set
\begin{equation}
    Y_i=f_d(X_i)+\frac{C_{0,d}Z_i}{\sqrt{U_i/5}}.
\end{equation}
The scalar $U_i$ is shared across all $d$ response coordinates.
Consequently,
\begin{equation}
    Y_i\mid X_i=x
    \sim t_{5,d}\bigl(f_d(x),S_{0,d}\bigr),
\end{equation}
where $t_{5,d}(\mu,S)$ denotes the multivariate Student-$t$
distribution with five degrees of freedom, location $\mu$, and
scale matrix $S$. Its conditional moments are
\begin{equation}
    \mathbb{E}[Y_i\mid X_i=x]=f_d(x),
    \qquad
    \operatorname{Cov}(Y_i\mid X_i=x)=\frac{5}{3}S_{0,d}.
\end{equation}
Thus, this setting introduces heavy-tailed residuals, while retaining
a fixed scale matrix. %Its noise covariance is larger than that ofCase~1 DPG3; no variance-matching rescaling is applied.

\subsection{Real-world datasets}
For the real-world experiments, we use the same datasets as in~\citet{Dheur2025-unified}\footnote{Parts of our implementation adapt open-source code from the \url{https://github.com/Vekteur/multi-output-conformal-regression} repository, 
which is distributed under the MIT License.}. Table~\ref{tab:datasets} summarizes the number of instances, input features, and output targets for each dataset.

\begin{table}[ht]
\centering
\caption{Summary of the real-world datasets.}
\label{tab:datasets}
\begin{tabular}{llccc}
\toprule
Source & Dataset & Nb. instances \(n\) & Nb. features \(p\) & Nb. targets \(d\) \\
\midrule
Cevid   & air        & 10000 & 15 & 6 \\
Cevid   & births1    & 10000 & 23 & 2 \\
Cevid   & births2    & 10000 & 24 & 4 \\
Camehl  & households & 7207  & 4  & 4 \\
Mulan   & rf1        & 9005  & 64 & 8 \\
Mulan   & rf2        & 9005  & 64 & 8 \\
Feldman & house      & 21613 & 14 & 2 \\
Feldman & bio        & 45730 & 8  & 2 \\
Wang    & taxi       & 50000 & 4  & 2 \\
\bottomrule
\end{tabular}
\end{table}
\paragraph{Data splitting and preprocessing.}
For each dataset and random seed, we split the data into training, calibration, and test sets. We cap the dataset sizes to 8000 samples. We first hold out \(20\%\) of the data for testing. The remaining \(80\%\) is then split into a training set and a calibration set, with the calibration set corresponding to \(25\%\) of this remaining data. This results in a \(60\%/20\%/20\%\) train/calibration/test split.

Both inputs and outputs are standardized using statistics computed on the training set only. The same transformations are then applied to the calibration and test sets. For PCP, we consider $L=200$ samples.

\subsection{Volume estimation.}
For hyperrectangular prediction sets, volumes are computed from the product of the coordinate-wise lengths. In our case all GH prediction sets of the scores considered have explicit closed form volume. For the PCP and log density prediction sets,  we estimate their volume using Monte Carlo integration, following the protocol of~\citet{Dheur2025-unified}.

For all bar plots, the bars report the mean value across random seeds, with the shaded \(\pm\) standard-deviation interval computed across seeds. For the \(\Delta_V\) metric, the displayed uncertainty is obtained by computing its the (empirical) standard error over the seeds.

\subsection{Additional Results}

\begin{figure}
    \centering
    \includegraphics[width=1\linewidth]{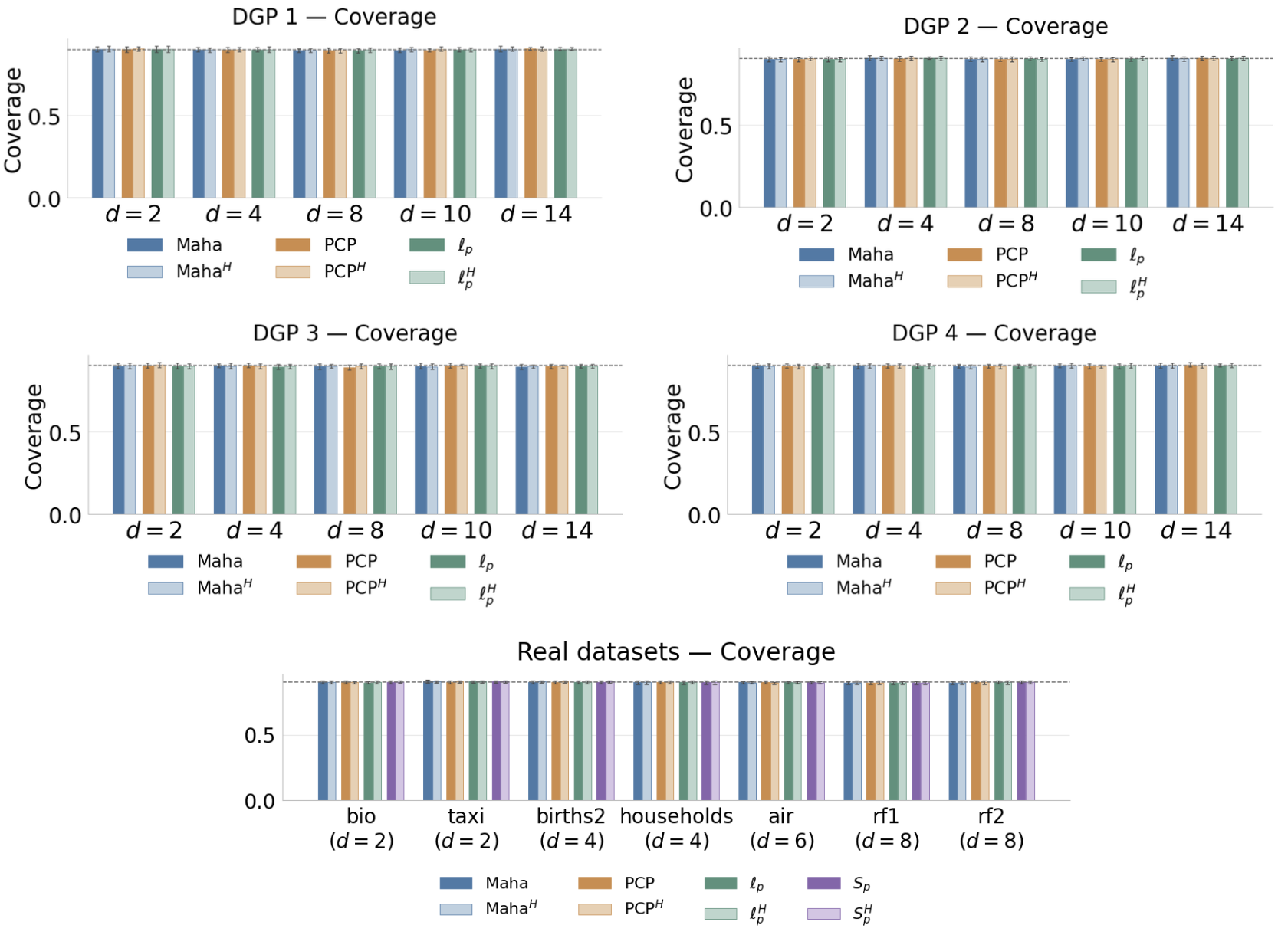}
    \caption{Mean empirical coverage for all DGPs and real-world datasets}
    \label{fig:all_cov_dgps}
\end{figure}

We report in Figure \ref{fig:all_cov_dgps}  the empirical coverage of all methods. As expected, all have coverage close to the nominal level $1-\alpha=0.9$. The finite-sample coverage guarantee follows directly from the SCP exchangeability argument of Theorem \ref{SCP_finite_sample_cov}.

\begin{figure}
    \centering
    \includegraphics[width=1\linewidth]{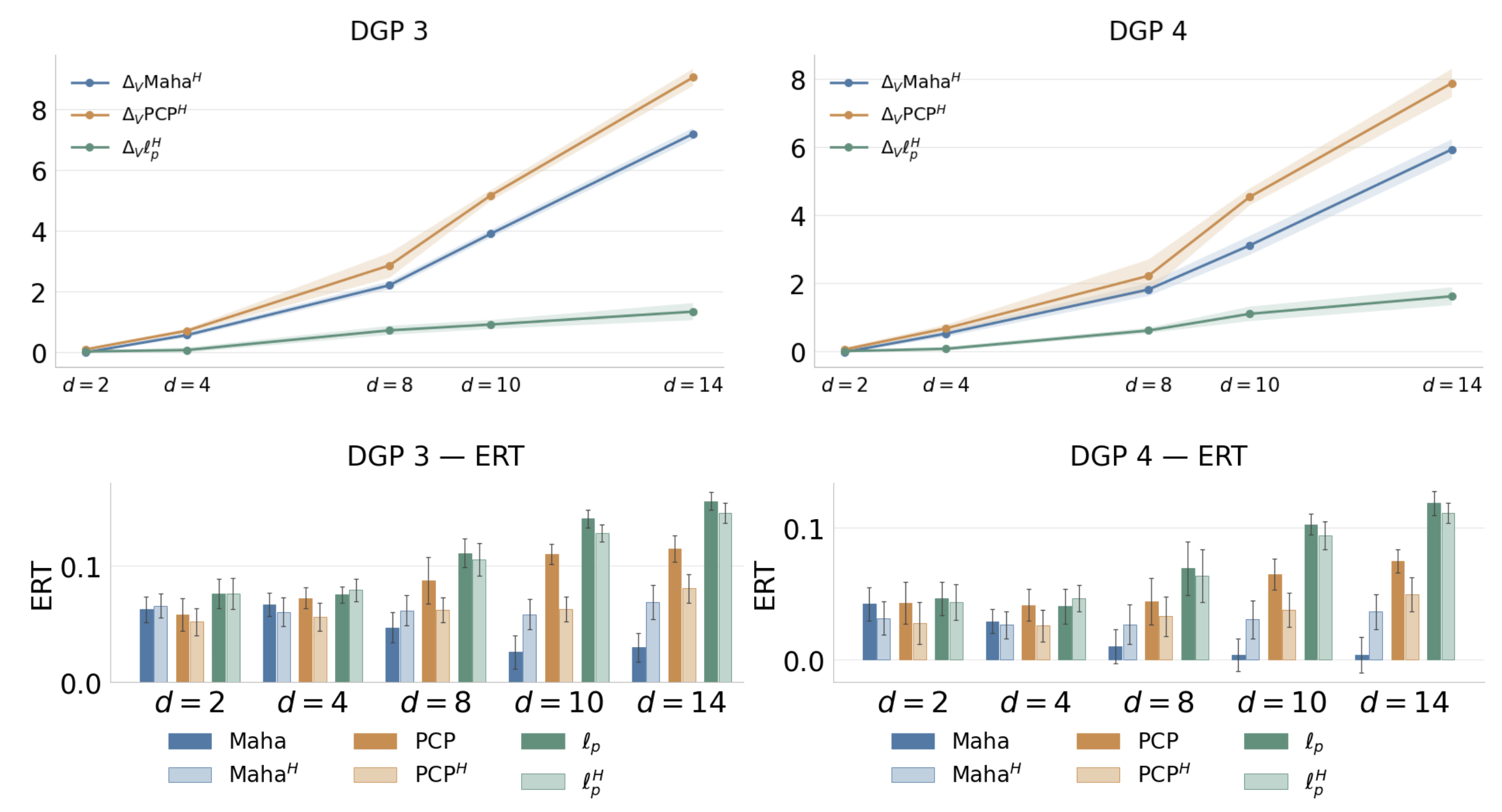}
    \caption{$\Delta_V$ and ERT metrics for all methods for DGPs 3 and 4.}
    \label{fig:Delta_V_DGP34}
\end{figure}

Regarding the log volume difference and the ERT metric for DGP 3 and DGP 4 in Figure \ref{fig:Delta_V_DGP34}, similar patterns are observed, where $\Delta_V$ increases with $d$, and the ERT of rectangularized prediction sets is usually slightly lower than that of the flexible sets.
%\subsection{Additional results:}

\section{Experiments with other baselines of the literature }
\label{app:experimental_details}

\paragraph{Methods.} We evaluate the performance of some hyperrectangular CP methods, i.e., with our notations and vocabulary, methods which yield generalized hyperrectangle prediction sets: $\mathrm{Maha}^H$, \textbf{M-CP}~\citep{MCP_zhou2024conformalized,Dheur2025-unified}, \textbf{CQR}~\citep{hyperectangle_CP}, \textbf{DCP}~\citep{dcp_mt}, \textbf{Max-Rank}~\citep{2025maxrank}.

In the real-world experiments, we use smaller values of \(R\), which were selected to ensure stable training: $R =1,  \text{for households, air, bio, and births2},
R=2,  \text{for rf1, taxi, and rf2}$.

\begin{figure}
    \centering
    \includegraphics[width=1\linewidth]{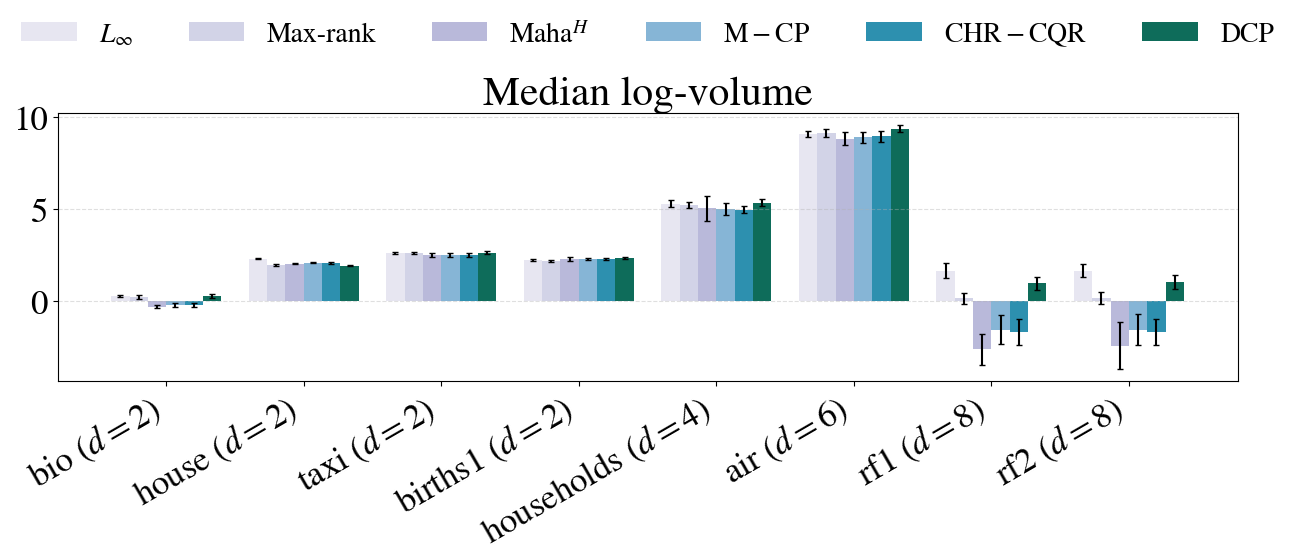}
    \caption{Median log-volume across real-world datasets of hyperrectangular CP methods of the literature.}
    \label{fig:real_results_CP_methods}
\end{figure}

The results are reported in Figure \ref{fig:real_results_CP_methods}. The relative efficiency of the hyperrectangular methods is somewhat similar across the datasets although it can be more pronounced across datasets. Several methods have nearby median logvolumes on datasets such as \texttt{taxi} and \texttt{births1}, whereas the differences are more pronounced on \texttt{rf1} and \texttt{rf2}. On these latter datasets, $\mathrm{Maha}^H$ achieves the lowest reported volume, while M-CP and CHR-CQR also produce smaller volumes than $L_\infty$, Max-rank, and DCP. 

%- scaling law 

%m ettre real dexp dan main paper not appendix

%%%%%%%%%%%%%%%%%%%%%%%%%%%%%%%%%%%%%%%%%%%%%%%%%%%%%%%%%%%%

\end{document}